\documentclass[letterpaper, preprint, paper, 11pt]{AAS}	
\usepackage[T1]{fontenc} 
\usepackage[utf8]{inputenc}
\usepackage[english]{babel}
\usepackage{amsmath,amssymb,amsfonts}
\usepackage{algorithm}
\usepackage{algorithmicx}
\usepackage{graphicx}
\usepackage{textcomp}
\usepackage{xcolor}
\usepackage{siunitx}
\usepackage{physics}
\usepackage{lineno}
\usepackage{subcaption}
\usepackage{booktabs}
\usepackage{tabularx}
\usepackage{threeparttable}
\usepackage{multirow}
\usepackage{longtable}
\usepackage{setspace}
\usepackage{float}
\usepackage{lipsum}
\usepackage[noend]{algpseudocode}
\usepackage{pdfpages}
\usepackage{enumitem}
\usepackage{graphicx}
\usepackage{color,soul}
\usepackage[normalem]{ulem}
\usepackage[version=4]{mhchem}
\usepackage{bm}

\usepackage{hyperref}

\usepackage{tikz}

\usetikzlibrary{positioning, arrows.meta, fit, backgrounds, shapes.geometric, arrows.meta, positioning, calc}
\definecolor{poscolor}{RGB}{54, 114, 181}
\definecolor{oricolor}{RGB}{124, 75, 141}
\definecolor{posecolor}{RGB}{84, 153, 94}
\usetikzlibrary{positioning, arrows.meta, fit, backgrounds, shapes.geometric, arrows.meta, positioning, calc}
\definecolor{imgcol}{RGB}{255,220,180}
\definecolor{ptcol}{RGB}{200,220,255}
\definecolor{dinocol}{RGB}{255,235,210}
\definecolor{dgcnncol}{RGB}{220,235,255}
\definecolor{matchcol}{RGB}{235,225,255}
\definecolor{sinkcol}{RGB}{220,245,230}
\definecolor{epnpcol}{RGB}{255,240,180}
\definecolor{reconcol}{RGB}{220, 235, 250} 
\definecolor{maskcol}{RGB}{200,200,200}

\tikzset{
  bigbox/.style={
    draw,
    rounded corners,
    align=center,
    minimum width=4.2cm,
    minimum height=2.2cm,
    font=\large
  },
  box/.style={
    draw,
    rounded corners,
    align=center,
    minimum width=3.4cm,
    minimum height=1.8cm,
    font=\normalsize
  },
  arrow/.style={
    ->,
    thick,
    line width=1.2pt
  }
}

\makeatletter
\renewcommand\paragraph{\@startsection{paragraph}{4}{\z@}%
  {-0.5ex \@plus -0.1ex \@minus -0.1ex}
  {-1em}
  {\normalfont\normalsize\bfseries}}    
\makeatother

\AtBeginDocument{%
  \setlength{\abovedisplayskip}{3pt}
  \setlength{\abovedisplayshortskip}{3pt}
  \setlength{\belowdisplayskip}{3pt} 
  \setlength{\belowdisplayshortskip}{3pt}
}

\AtBeginDocument{%
  \setlength{\textfloatsep}{8pt plus 2pt minus 2pt} 
  
  \setlength{\intextsep}{8pt plus 2pt minus 2pt} 
  
  \setlength{\floatsep}{6pt plus 2pt minus 2pt} 
  
  \setlength{\abovecaptionskip}{3pt} 
  \setlength{\belowcaptionskip}{0pt}

  \setlength{\dbltextfloatsep}{10pt plus 2pt minus 4pt}
}

\PaperNumber{26-790}

\makeatletter
\renewcommand*{\@fnsymbol}[1]{\ensuremath{\ifcase#1\or 1\or 2\or 3\or 4\or 5\or 6\or 7\or 8\or 9\or 10\or 11\or 12\or 13\or 14\or 15\else\@ctrerr\fi}}
\makeatother

\makeatletter
\def\blfootnote{\xdef\@thefnmark{}\@footnotetext}
\makeatother

\begin{document}


\title{DreamSat-Pose: Spacecraft Pose Estimation from Single-View 3D Reconstructions and Learned 2D-3D Feature Matching}


\author{
Josiane Uwumukiza \thanks{Undergraduate Student, Wellesley College, Wellesley, MA 02481, USA, \textit{ju100@wellesley.edu.}},
\ Jocelyn Zhao \thanks{Undergraduate Student, Massachusetts Institute of Technology, Cambridge, MA 02139, USA, \textit{jjz300@mit.edu.}},
\ Giovanni Lavezzi\thanks{Research Scientist, Massachusetts Institute of Technology, Cambridge, MA 02139, USA, \textit{glavezzi@mit.edu.}}, 
\ Giacomo Battaglia \thanks{Ph.D. Student, Politecnico di Milano, Milan, Italy, \textit{giacomo.battaglia@polimi.it.}},
\ Paolo Panicucci \thanks{Assistant Professor, Politecnico di Milano, Milan, Italy, \textit{paolo.panicucci@polimi.it.}},
\ Minduli C. Wijayatunga \thanks{Assistant Professor, University of Illinois Urbana-Champaign, Urbana, IL 61801, USA, \textit{minduli@illinois.edu.}},
\ Victor Rodriguez-Fernandez \thanks{Associate Professor, Universidad Politécnica de Madrid, Madrid, Spain, \textit{victor.rfernandez@upm.es.}},
\ and Richard Linares\thanks{Associate Professor, Massachusetts Institute of Technology, Cambridge, MA 02139, USA, \textit{linaresr@mit.edu}}
}

\maketitle

\begin{abstract}
6-DoF pose estimation is a critical task in autonomous rendezvous and proximity operations. In the case of an unknown target, this task is especially challenging because it must be paired with the reconstruction of the target shape model. In this article, we propose a novel framework for single-shot shape and pose estimation of unknown spacecraft objects. Given a single image, we first reconstruct a 3D shape model of the target, then estimate the relative six-degree-of-freedom pose by learning dense 2D-3D correspondences. The image features are extracted using a frozen DINOv3 vision transformer, while the geometric features are computed from the reconstructed point cloud using a trainable dynamic graph convolutional neural network encoder. A dual-stream transformer matcher refines descriptors through alternating self- and cross-attention, producing soft correspondences that are passed to a Perspective-$n$-Point solver for pose recovery. We evaluate the method on the SPE3R dataset and consider FoundationPose as a representative baseline for current state-of-the-art capabilities. Results show reliable pose estimates achieving a mean pointing error of \(0.157^\circ\) using only a single image and reconstructed geometry, demonstrating strong generalization to unseen spacecraft.

\end{abstract}

\section{Introduction}

6-degree-of-freedom (6-DoF) pose estimation is a critical capability for space robotic missions, including on-orbit docking, inspection, navigation, and debris removal, where perception outputs must support real-time guidance and control \cite{wijayatunga2026cortex}. In these scenarios, spacecraft must often estimate the pose of an uncooperative target using extremely limited sensing. In particular, monocular cameras are frequently the only available sensors, providing a single RGB (Red-Green-Blue) image without reliable depth or scale information, making accurate pose estimation challenging. Classical Perspective-$n$-Point (PnP) pipelines \cite{PAULY2023339} and recent deep learning-based methods, particularly Convolutional Neural Networks (CNNs) \cite{bates2025removing, aerospaceServadio}, frequently struggle to generalize to unseen spacecraft \cite{chen2019satellite, 10225381}, limiting their operational utility.

To address these challenges, we leverage recent advances in representation learning. DINOv3 \cite{simeoni2025dinov3} provides viewpoint-consistent visual descriptors, while Dynamic Graph CNN (DGCNN) \cite{dgcnn} extracts expressive features from 3D point clouds by modeling local geometric relationships. While 2D matchers like SuperGlue \cite{Sarlin2020SuperGlue} or registration tools like MinCD-PnP \cite{An2025MinCDPnP} have made progress, they remain limited by a lack of direct 2D-3D matching or a reliance on metrically scaled Computer Aided Design (CAD) models often obtained from depth sensors or structure-from-motion, which are not available in our setting. Alternatively, generative frameworks like DreamSat \cite{dreamsat, dreamsat2} use diffusion priors to reconstruct 3D geometries from single observations. 

In this work, we propose a single-view, correspondence-based framework for 6-DoF pose estimation that leverages reconstructed object geometry without requiring depth or multi-view supervision. Building on our previous work, DreamSat \cite{dreamsat,dreamsat2}, which enables single-view 3D reconstruction of spacecraft, we use the reconstructed structure as a geometric prior to guide pose estimation. This reformulates the problem from direct pose regression into a structured correspondence problem between 2D image features and 3D model points. We extract robust 2D features and keypoints using a frozen vision transformer (DINOv3) \cite{simeoni2025dinov3} from the image, train a graph-based network (DGCNN) \cite{dgcnn} to encode the 3D CAD, and learn soft 2D-3D correspondences via a lightweight transformer with self- and cross-attention. These correspondences are then used within a classical PnP solver to recover the pose of the object. 
To evaluate the proposed framework, we utilize the SPE3R dataset \cite{spe3r_data,park2024rapid}, including 64 unique spacecraft models. We also consider NVIDIA’s FoundationPose \cite{foundationposewen2024}, a generalist 6D pose estimator capable of zero-shot generalization to unseen objects, as a representative baseline for current state-of-the-art capabilities. It must be noted that FoundationPose requires depth maps as inputs, whereas our methodology operates strictly on monocular RGB data; as such, FoundationPose serves as a best-case performance benchmark rather than a direct equivalent. We employ two primary experimental configurations: a ground-truth-based assessment to establish a performance upper bound, and a reconstruction-based assessment to evaluate the system’s utility for unknown targets. This analysis investigates whether generative 3D proxies can effectively substitute for a-priori CAD models in autonomous proximity operations. By combining modern representation learning with geometric constraints, our method achieves viable pose estimation from a single RGB image. 

The paper is organized as follows. First, we introduce the single-view correspondence-based framework for 6-DoF pose estimation, which is called DreamSat-Pose. Then, we describe the mission scenario, dataset and evaluation metrics. Subsequently, we discuss the experimental design and results of the proposed pose estimation pipeline. Lastly, we provide final remarks to conclude the paper.



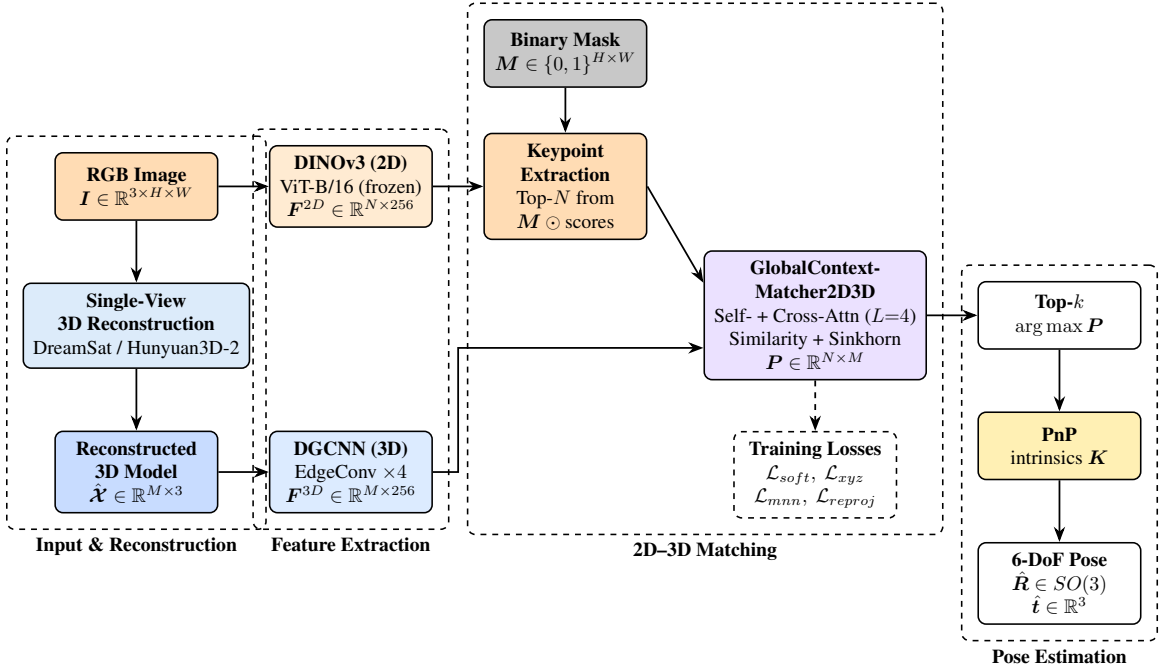
\begin{figure}[h]
\centering
\resizebox{\textwidth}{!}{%
\begin{tikzpicture}[
    font=\small,
    node distance=1.1cm and 0.9cm,
    every node/.style={align=center},
    box/.style={draw, rounded corners, inner sep=5pt, line width=0.8pt,
                minimum width=2.9cm, minimum height=1.2cm},
    arrow/.style={-{Stealth[scale=1.1]}, line width=0.9pt},
    grp/.style={draw, dashed, rounded corners, line width=0.8pt, inner sep=8pt},
    grplabel/.style={font=\small\bfseries}
]
\node[box, fill=imgcol] (img)
  {\textbf{RGB Image}\\
   $\bm{I} \in \mathbb{R}^{3\times H \times W}$};
\node[box, fill=reconcol, below=of img] (recon)
  {\textbf{Single-View}\\ \textbf{3D Reconstruction}\\
   DreamSat / Hunyuan3D-2};
\node[box, fill=ptcol, below=of recon] (cad)
  {\textbf{Reconstructed}\\ \textbf{3D Model}\\
   $\hat{\bm{\mathcal{X}}} \in \mathbb{R}^{M \times 3}$};

\node[box, fill=dinocol, right=of img] (dino)
  {\textbf{DINOv3 (2D)}\\
   ViT-B/16 (frozen)\\
   $\bm{F}^{2D} \in \mathbb{R}^{N\times256}$};
\node[box, fill=dgcnncol, right=of cad] (dgcnn)
  {\textbf{DGCNN (3D)}\\
   EdgeConv $\times 4$\\
   $\bm{F}^{3D} \in \mathbb{R}^{M\times256}$};

\node[box, fill=imgcol, right=of dino] (kp)
  {\textbf{Keypoint}\\ \textbf{Extraction}\\
   Top-$N$ from\\ $\bm{M} \odot \text{scores}$};
\node[box, fill=maskcol, above=0.8cm of kp] (mask)
  {\textbf{Binary Mask}\\
   $\bm{M} \in \{0,1\}^{H \times W}$};

\node[box, fill=matchcol, right=1.0cm of kp, yshift=-2.3cm] (matcher)
  {\textbf{GlobalContext-}\\ \textbf{Matcher2D3D}\\
   Self- + Cross-Attn ($L{=}4$)\\
   Similarity + Sinkhorn\\
   $\bm{P} \in \mathbb{R}^{N\times M}$};
\node[box, dashed, below=0.9cm of matcher] (loss)
  {\textbf{Training Losses}\\
   $\mathcal{L}_{soft},\ \mathcal{L}_{xyz}$\\
   $\mathcal{L}_{mnn},\ \mathcal{L}_{reproj}$};

\node[box, right=of matcher] (topk)
  {\textbf{Top-$k$}\\
   $\arg\max \bm{P}$};
\node[box, fill=epnpcol, below=of topk] (epnp)
  {\textbf{PnP}\\
   intrinsics $\bm{K}$};
\node[box, below=of epnp] (pose)
  {\textbf{6-DoF Pose}\\
   $\hat{\bm{R}} \in SO(3)$\\
   $\hat{\bm{t}} \in \mathbb{R}^3$};

\draw[arrow] (img) -- (recon);
\draw[arrow] (recon) -- (cad);
\draw[arrow] (img) -- (dino);
\draw[arrow] (cad) -- (dgcnn);
\draw[arrow] (dino) -- (kp);
\draw[arrow] (mask) -- (kp);
\draw[arrow] (kp.east) -- ($(matcher.north west)!0.25!(matcher.south west)$);
\draw[arrow] (dgcnn.east) -- ++(0.45,0) |- ($(matcher.north west)!0.75!(matcher.south west)$);
\draw[arrow, dashed] (matcher) -- (loss);
\draw[arrow] (matcher) -- (topk);
\draw[arrow] (topk) -- (epnp);
\draw[arrow] (epnp) -- (pose);

\begin{scope}[on background layer]
\node[grp, fit=(img)(recon)(cad),
      label={[grplabel]below:Input \& Reconstruction}] {};
\node[grp, fit=(dino)(dgcnn),
      label={[grplabel]below:Feature Extraction}] {};
\node[grp, fit=(mask)(kp)(matcher)(loss),
      label={[grplabel]below:2D--3D Matching}] {};
\node[grp, fit=(topk)(epnp)(pose),
      label={[grplabel]below:Pose Estimation}] {};
\end{scope}
\end{tikzpicture}%
}
\caption{Overview of the DreamSat-Pose pipeline}
\label{fig:architecture}
\end{figure}

\section{DreamSat-Pose} \label{sec:s2}
Our pipeline consists of four main components: (a) 3D reconstruction from a single RGB image of the space object, (b) feature extraction on the image and 3D reconstructed model, (c) feature matching to produce 2D-3D correspondences, and (d) pose solving to estimate the camera pose from matches. Figure~\ref{fig:architecture} illustrates the flow of data through these stages. An RGB image is given as input to DreamSat to obtain a 3D reconstructed model. The same image is encoded using a frozen DINOv3 vision transformer to extract 2D features, while the reconstructed CAD model is processed by a DGCNN to obtain 3D descriptors. A transformer-based cross-attention module produces a soft correspondence matches. Top-$k$ correspondences are used in an EPnP solver \cite{lepetit2009ep} to estimate pose.

\subsection{Input Image Processing Techniques}
The input images from the SPE3R dataset are utilized at their native resolution $256 \times 256$ for the majority of the pipeline. However, we observed that when a spacecraft occupies only a small portion of the total image area the pose estimation performance degrades significantly. In these instances, the effective resolution of the target is insufficient for the DINOv3 encoder to extract discriminative visual descriptors, leading to unstable 2D-3D correspondences. To mitigate this, we employ a bounding-box-aware cropping and upscaling strategy. First, we utilize the segmentation mask to identify the tightest bounding box surrounding the detected spacecraft. This region is then cropped from the original frame and rescaled back to the $256 \times 256$ input size using bilinear interpolation. The corresponding segmentation mask is resized using nearest-neighbor interpolation to preserve binary foreground/background labels. This process maximizes the pixel density of the satellite's structure within the input window and provides the feature matcher with a more consistent representation, thereby improving the robustness of the PnP solver. Future work will focus on improving the quality of correspondences, which, despite the cropping and upscaling strategy, remain a challenge. We will also explore image enhancements such as super-resolution and denoising and higher resolution images to further improve feature matching and reconstruction fidelity.

\subsection{Single-View 3D Reconstruction}
DreamSat integrates and benchmarks multiple state-of-the-art single-view generative 3D reconstruction models as described in our previous work \cite{dreamsat, dreamsat2}. In this work, we adopt Hunyuan-3D-2.0 \cite{hunyuan3d22025tencent} due to its strong performance across both 2D perceptual and 3D geometric metrics on custom spacecraft and asteroid datasets, making it well-suited for downstream pose estimation tasks. As a result, DreamSat-Pose leverages Hunyuan-3D-2.0 reconstructions as the primary geometric representation in the proposed pose estimation pipeline. By reconstructing spacecraft geometry from a single observation, the proposed framework enables pose estimation in scenarios where traditional model-based approaches are not feasible and no additional observations and depth information are available. 

\subsection{Image Feature Extraction} 
To enable generalization to unseen targets, a frozen DINOv3 ViT-B/16 \cite{simeoni2025dinov3} is used as the 2D feature extractor and keypoint selector. Given a $256\times256$ RGB image, the model produces a dense grid of patch descriptors (size $16\times16$ with patch size 16), where each descriptor is 768-dimensional. These descriptors are projected to 256 dimensions via a linear layer to reduce memory and align with the 3D feature space produced by the DGCNN encoder. In addition to descriptors, a lightweight score head produces a scalar saliency for each patch, highlighting informative regions such as edges and textured areas while remaining robust to harsh orbital lighting. Keypoints are selected by thresholding this score, yielding a set of candidate 2D feature points. A foreground mask is applied to ensure that the selected key points are on the object. 

\subsection{3D Feature Extraction} After 3D reconstruction, 1024 points are uniformly sampled from the surface mesh and normalized through centering and scaling to ensure a geometry-independent representation, serving as our candidate 3D keypoints. These points are fed into the DGCNN network, which constructs a $k$-NN graph on the points ($k=20$) neighbors and applies a series of EdgeConv layers. EdgeConv is a graph convolution operator that learns local geometric features by computing edge representations between each point and its neighbors, allowing the network to capture both the properties of individual points and their local spatial relationships. In an EdgeConv, for each point, a small neural network processes the concatenation of the point's feature and the relative feature of a neighbor, thereby encoding geometric relationships. The descriptors from four successive layers are concatenated to aggregate local surface curvature with global object context, resulting in a 256-dimensional embedding. After each EdgeConv, a max-pooling over the neighbors produces an updated feature for each point. We initialize the DGCNN randomly and train it jointly with the matcher. This allows the 3D features to adapt to the distribution of our object models. The DGCNN is optimized using the matcher's alignment and reprojection losses to co-adapt 3D geometric descriptors into a shared latent space with the 2D image features. 

\subsection{Cross-Modal Matching Network} 
\begin{figure}[h]
\centering
\resizebox{0.4\textheight}{!}{%
\begin{tikzpicture}[
    scale=1, every node/.style={scale=1},
    node distance=0.4cm and 0.4cm,
    box/.style={rectangle, draw, thick, minimum width=2.0cm, minimum height=0.5cm, align=center, font=\small\sffamily},
    op/.style={circle, draw, thick, inner sep=0pt, minimum size=3.5mm, font=\small},
    res_node/.style={box, fill=blue!10},
    stream2d/.style={box, fill=cyan!15},
    stream3d/.style={box, fill=orange!15},
    shared/.style={box, fill=gray!10},
    arrow/.style={-Stealth, thick}
]

\node[stream2d] (f2d) {2D Features};
\node[stream2d, right=0.2cm of f2d] (c2d) {2D Coords};
\node[op, below=0.4cm of $(f2d.south)!0.5!(c2d.south)$] (add2d) {+};

\draw[arrow] (f2d) |- (add2d);
\draw[arrow] (c2d) |- (add2d);

\node[res_node, below=0.4cm of add2d] (self2d) {Self-Attention};
\node[res_node, below=0.3cm of self2d] (cross2d) {Cross-Attention};

\node[stream3d, right=0.8cm of c2d] (f3d) {3D Features};
\node[stream3d, right=0.2cm of f3d] (c3d) {3D Coords};
\node[op, below=0.4cm of $(f3d.south)!0.5!(c3d.south)$] (add3d) {+};

\draw[arrow] (f3d) |- (add3d);
\draw[arrow] (c3d) |- (add3d);

\node[res_node, below=0.4cm of add3d] (self3d) {Self-Attention};
\node[res_node, below=0.3cm of self3d] (cross3d) {Cross-Attention};

\node[draw, dashed, gray, inner sep=0.15cm, fit=(self2d) (cross3d), label={[gray]right:$\times 4$}] (block) {};

\draw[arrow, <->, dashed, color=red!70] (cross2d.east) -- (cross3d.west) node[midway, above, text=black] {Context};

\node[shared, below=0.6cm of cross2d] (norm2d) {$L_2$ Norm};
\node[shared, below=0.6cm of cross3d] (norm3d) {$L_2$ Norm};

\node[op, below=0.4cm of $(norm2d.south)!0.5!(norm3d.south)$] (bmm) {$\otimes$};
\node[shared, below=0.3cm of bmm, minimum width=4cm] (sinkhorn) {Sinkhorn Normalization};
\node[box, fill=green!15, below=0.3cm of sinkhorn, minimum width=4cm] (out) {Correspondence Matrix $P$};
\draw[arrow] (add2d) -- (self2d);
\draw[arrow] (self2d) -- (cross2d);
\draw[arrow] (cross2d) -- (norm2d);
\draw[arrow] (add3d) -- (self3d);
\draw[arrow] (self3d) -- (cross3d);
\draw[arrow] (cross3d) -- (norm3d);
\draw[arrow] (norm2d) |- (bmm);
\draw[arrow] (norm3d) |- (bmm);
\draw[arrow] (bmm) -- (sinkhorn);
\draw[arrow] (sinkhorn) -- (out);

\end{tikzpicture}
}
\caption{Dual-stream transformer architecture for 2D–3D correspondence matching. 2D image features and 3D geometric features are refined through alternating self- and cross-attention layers, followed by Sinkhorn normalization to produce the soft correspondence matrix $P$.
}
\label{fig:matching}
\end{figure}
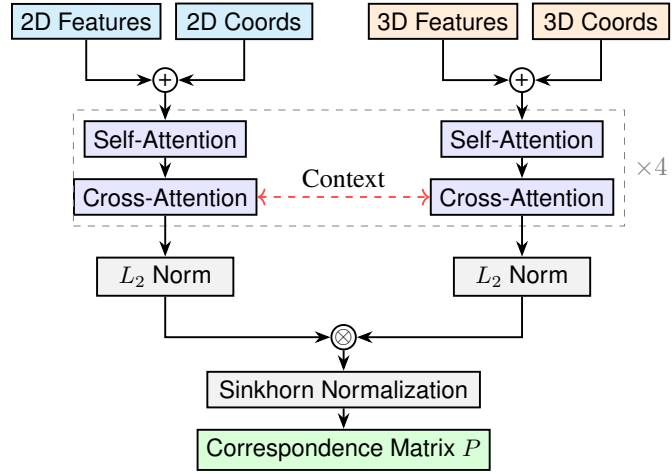

The 2D-3D matching module takes as input the set of 2D keypoint descriptors $\{f_i^{\text{2D}}\}$ and 3D point descriptors $\{f_j^{\text{3D}}\}$ (of size $N$ and $M$ respectively), and outputs a soft correspondence matrix $P \in \mathbb{R}^{N \times M}$, as shown in Fig. \ref{fig:matching}. The architecture follows a Transformer-based dual encoder-decoder design, inspired by SuperGlue \cite{Sarlin2020SuperGlue}. Both descriptor sets are projected into a common 256-dimensional space via linear layers, then augmented with positional encodings: normalized 2D coordinates $(x, y \in [0,1])$ and normalized 3D coordinates (XYZ) are each passed through small Multilayer Perceptrons (MLPs) and added to their respective descriptors. Separate self-attention layers perform intra-modality refinement, followed by interleaved cross-attention layers where queries from one modality attend to keys and values from the other, enabling correspondence reasoning across modalities. The cosine similarity matrix $S \in \mathbb{R}^{N \times M}$ is computed using the dot product of the $L_2$-normalized features:
\begin{equation}
    S_{ij} = \langle \hat{f}_i^{\text{2D}}, \hat{f}_j^{\text{3D}} \rangle
\end{equation}
Sinkhorn normalization\cite{Cuturi2013Sinkhorn} is then applied to $S$, yielding a doubly-stochastic matrix $P$ that enforces one-to-one matching and remains differentiable during training. At inference, the top match per 2D point is selected as $\arg\max_j P_{ij}$, with matches below a confidence threshold discarded.

\subsection{Loss Functions} 

We introduce a supervised geometric objective that leverages known camera poses to generate correspondence targets. Rather than direct parameter regression, the ground-truth (GT) rotation $R_{gt}$, translation $t_{gt}$, and camera intrinsics $K$ are used to project each sampled 3D point $X_j \in \mathbb{R}^3$ into the image plane as $\tilde{u}_j = \pi(K(R_{gt}X_j + t_{gt}))$, where $\pi(\cdot)$ denotes perspective projection. For each selected 2D keypoint $u_i \in \mathbb{R}^2$, a soft GT correspondence distribution $P^{gt}_{ij}$ is constructed by applying a local softmax over Gaussian-weighted Euclidean distances $d_{ij} = \|u_i - \tilde{u}_j\|_2$ within a top-$k$ neighborhood. This soft formulation accounts for the inherent ambiguity of the correspondence problem, where multiple 3D points may project near a single 2D keypoint due to sparse sampling or local symmetry, thereby avoiding brittle hard assignments \cite{li2020correspondence}. The predicted soft assignment matrix $P$ is then optimized via the soft correspondence loss:
\begin{equation}
\mathcal{L}_{\text{soft}} = -\frac{1}{|\mathcal{V}|} \sum_{i \in \mathcal{V}} \sum_{j=1}^{M} P^{gt}_{ij} \log P_{ij}
\end{equation}
where $\mathcal{V}$ denotes the set of valid 2D keypoints and $M$ is the total number of 3D points. To provide a strong geometric signal to the DGCNN, we employ an expected-XYZ loss \cite{brachmann2018learning} to align the predicted expected 3D point $\hat{X}_i = \sum_{j=1}^{M} P_{ij} X_j$ with the target 3D point $X_i^{gt} = \sum_{j=1}^{M} P^{gt}_{ij} X_j$:
\begin{equation}
\mathcal{L}_{\text{xyz}} = \frac{1}{|\mathcal{V}|} \sum_{i \in \mathcal{V}} \left\| \hat{X}_i - X_i^{gt} \right\|_2^2
\end{equation}
Direct pixel-space consistency is enforced by the reprojection loss \cite{bates2025removing,park2019speed}, which minimizes the distance between the observed 2D keypoint $u_i$ and the reprojection of the predicted soft match $\hat{X}_i$:
\begin{equation}
\mathcal{L}_{\text{reproj}} = \frac{1}{|\mathcal{V}|} \sum_{i \in \mathcal{V}} \left\| \pi(K(R_{gt}\hat{X}_i + t_{gt})) - u_i \right\|_2
\end{equation}
The assignment matrix is further regularized by a mutual nearest-neighbor loss $\mathcal{L}_{\text{mnn}}$ to ensure bidirectional matching coherence \cite{Sarlin2020SuperGlue, tyszkiewicz2020disk}. The total supervised objective is a weighted combination of these terms:
\begin{equation}
\mathcal{L}_{\text{sup}} = \lambda_{\text{soft}} \mathcal{L}_{\text{soft}} + \lambda_{\text{xyz}} \mathcal{L}_{\text{xyz}} + \lambda_{\text{mnn}} \mathcal{L}_{\text{mnn}} + \lambda_{\text{reproj}} \mathcal{L}_{\text{reproj}}
\end{equation}
using weights $\lambda_{\text{soft}}=1.0$, $\lambda_{\text{xyz}}=1.0$, $\lambda_{\text{mnn}}=0.5$, and $\lambda_{\text{reproj}}=0.01$. This formulation allows the matcher and DGCNN to co-adapt into a shared latent space, stabilized by an annealing curriculum that gradually sharpens the Gaussian bandwidth $\sigma$ and neighborhood size $k$ over time.

\subsection{Pose Calculation} 
\begin{figure}[!t]
    \centering
    \captionsetup{font=small}

    \begin{subfigure}[b]{0.40\linewidth}
        \centering
        \includegraphics[width=\linewidth]{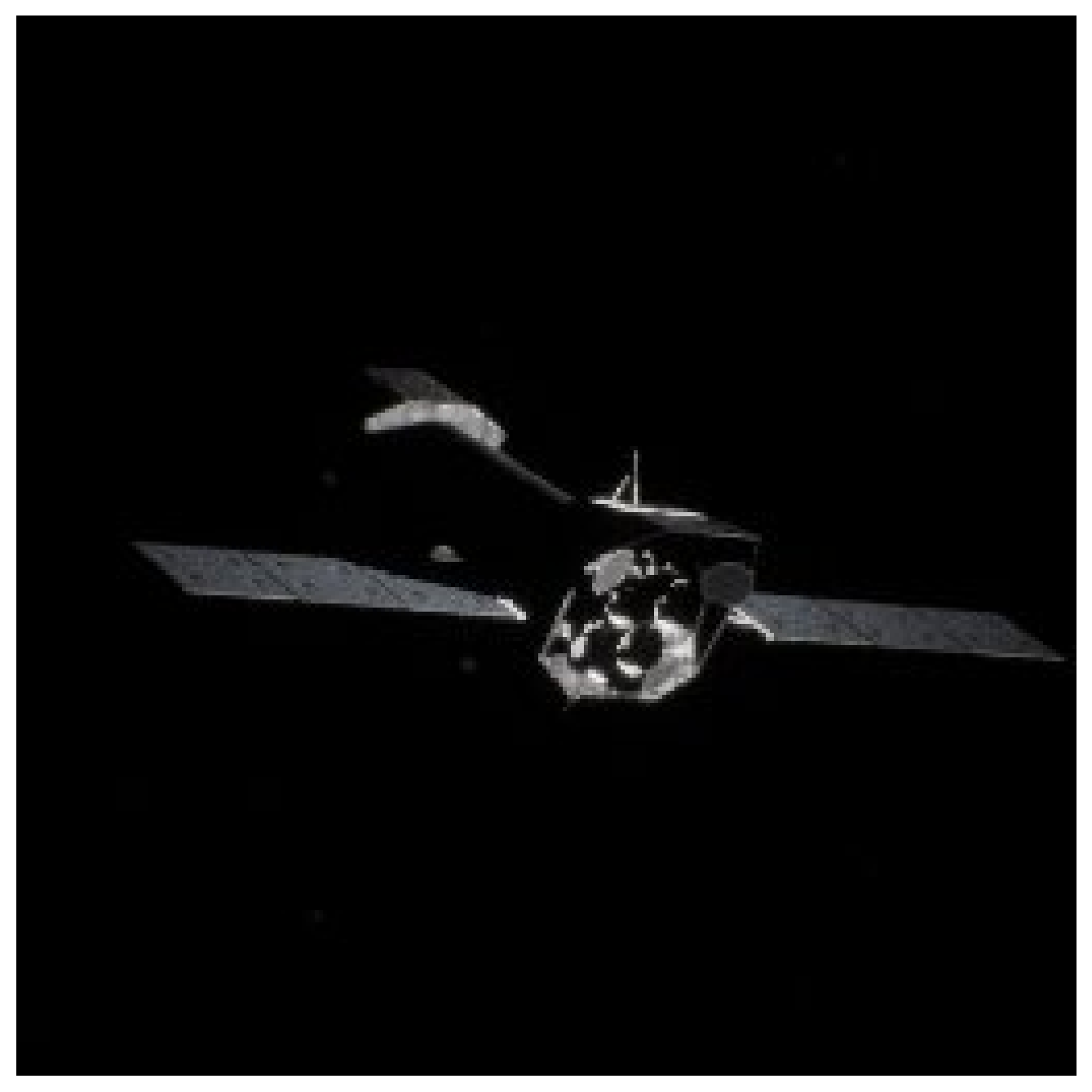}
        \caption{Input image.}
        \label{fig:input_img}
    \end{subfigure}
    \hspace{0.04\linewidth}
    \begin{subfigure}[b]{0.40\linewidth}
        \centering
        \includegraphics[width=\linewidth]{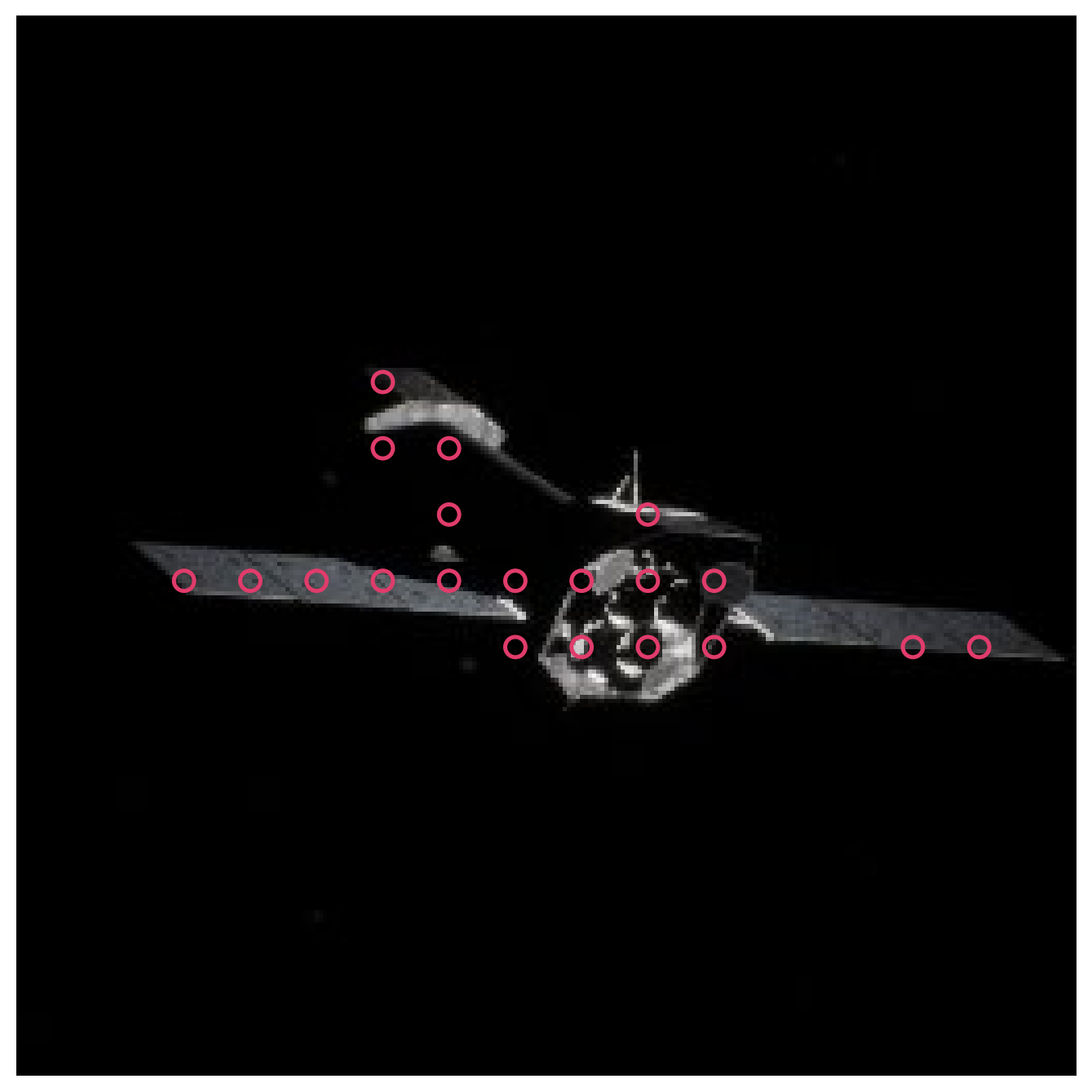}
        \caption{DINOv3 keypoints.}
        \label{fig:dino_kp}
    \end{subfigure}

    \par\smallskip

    \begin{subfigure}[b]{0.40\linewidth}
        \centering
        \includegraphics[width=\linewidth]{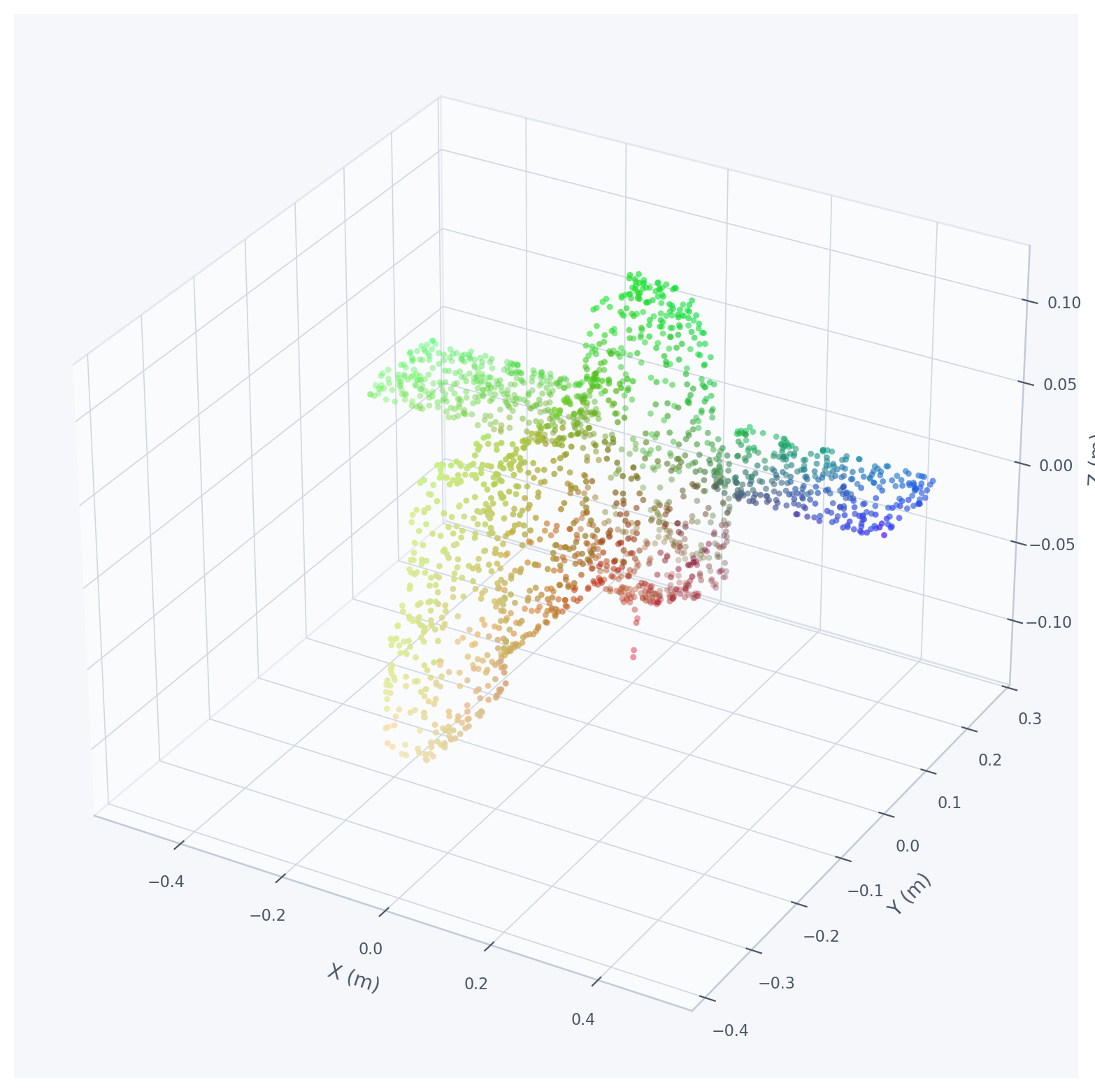}
        \caption{DGCNN 3D features.}
        \label{fig:3d_feats}
    \end{subfigure}
    \hspace{0.04\linewidth}
    \begin{subfigure}[b]{0.40\linewidth}
        \centering
        \includegraphics[width=\linewidth]{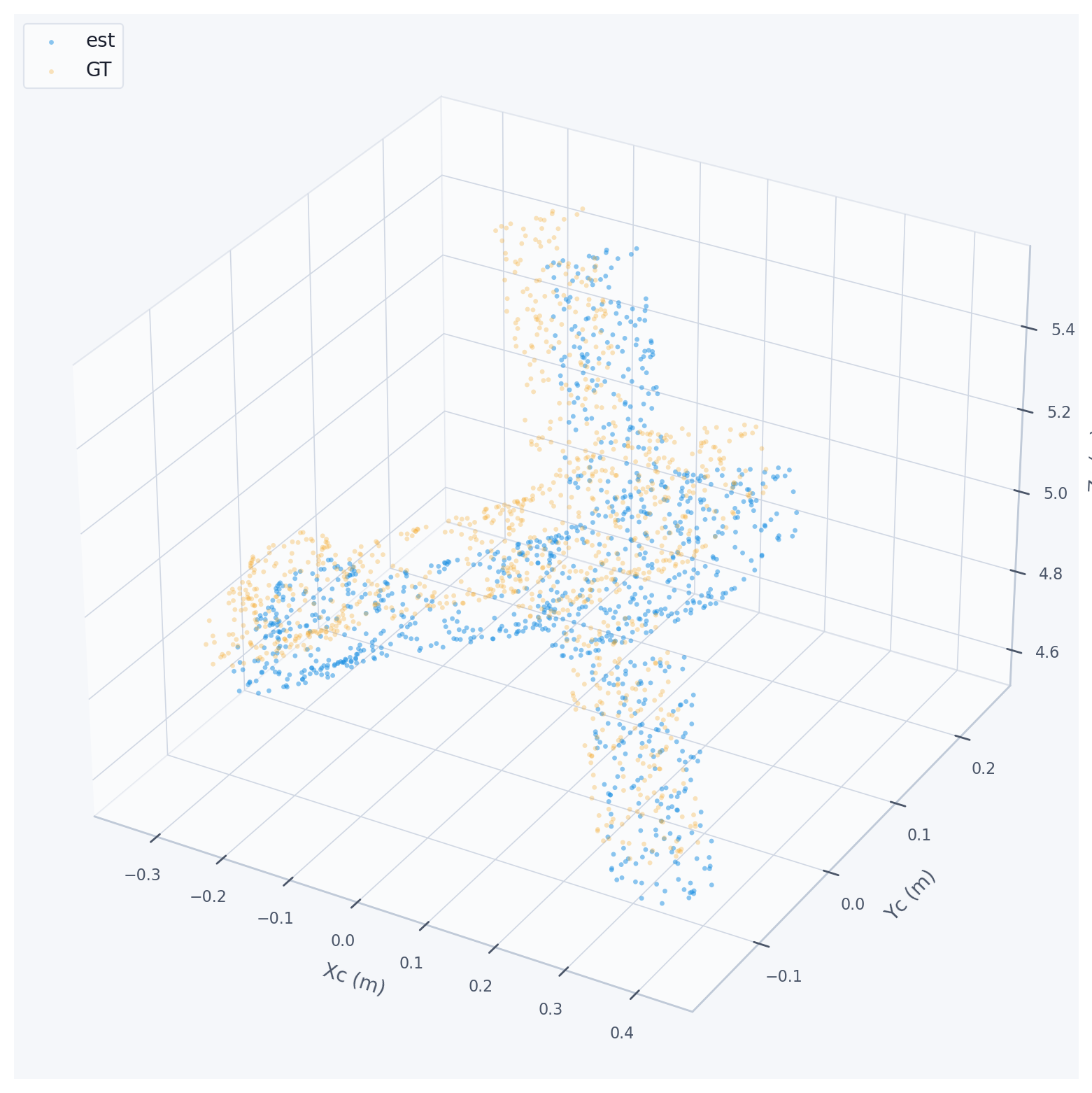}
        \caption{Estimated pose alignment.}
        \label{fig:pose_aln}
    \end{subfigure}

     \caption{
        Qualitative results of the proposed DreamSat-Pose pipeline.
        (a) Input RGB image.
        (b) Keypoints selected by DINOv3, which are distributed across geometrically informative regions.
        (c) Learned 3D feature representation of the reconstructed object using DGCNN.
        (d) Estimated pose alignment in 3D space, showing agreement between the reconstructed model (blue) and ground truth (orange).
     }
    \label{fig:qualitative}
\end{figure}

To translate learned correspondences into a 6-DoF pose $(R, t)$, we employ a robust geometric pipeline. Discrete matches are first filtered by the confidence scores in $P$, requiring a minimum of six points to avoid underdetermined solutions. To handle outliers arising from geometric symmetries or reconstruction noise, an initial pose is estimated using the Efficient PnP (EPnP) algorithm \cite{lepetit2009ep} within a RANSAC (RANdom SAmple Consensus) framework of 200 iterations. This estimate is subsequently refined through iterative nonlinear Levenberg-Marquardt optimization to minimize the total reprojection error of the identified inlier set. This hybrid approach combines the generalization of learned features with the interpretability and precision of classical geometric solvers. Figure \ref{fig:qualitative} shows the qualitative results of the proposed DreamSat-Pose pipeline.

\section{Experimental Setup} \label{sec:s3}

\subsection{Mission Scenario and Assumptions}
The proposed methodology aims to estimate the 6-DoF pose of an unknown spacecraft from a single RGB image without access to GT geometry, GT CAD models, or multi-view observations. We assume monocular sensing, unknown spacecraft 3D model, and no depth measurements. We assume non-cooperative targets, with the absence of fiducial markers or predefined landmarks. These assumptions define a challenging regime for pose estimation. Under these constraints, the system must rely on learned geometric reasoning. These constraints also influence design decisions throughout the pipeline. For example, reconstruction is performed from a single image rather than multi-view geometry. Similarly, learned descriptors are used instead of handcrafted features to improve robustness under varying conditions. With these constraints, this formulation closely reflects realistic operational conditions in space autonomy systems.

\subsection{Dataset}
The Spacecraft Pose Estimation and 3D Reconstruction (SPE3R) dataset \cite{spe3r_data} is an open-source satellite image dataset developed at Stanford's Space Rendezvous Laboratory (SLAB) \cite{park2024rapid}. SPE3R comprises 64 unique satellite models, and each satellite model is characterized by 1,000 synthetic images and their corresponding binary segmentation masks, GT pose labels (orientation and position) and 3D point clouds. The synthetic images in SPE3R are of size $256\times256$ pixels. These images depict a satellite against either a black or Earth background, with added illumination features such as harsh lighting and shadows. 

\subsection{Data Splitting}
For comparison, we adopt the same dataset split as used in the literature \cite{bates2025removing}. The dataset is divided into 57 training spacecraft and 7 unseen test spacecraft. Each spacecraft contains 1000 images with unique poses. This results in 7000 images. For each training object, the images are split 80:20 into training and validation subsets such that the validation set is a group of unseen images of a known spacecraft. Both training and inference are performed on a workstation equipped with NVIDIA RTX 4090 hardware.

\subsection{Reconstruction Metrics}
To evaluate the fidelity of the 3D reconstructions, we employ 3D geometric metrics \cite{dreamsat2}. Specifically, the structural accuracy is assessed by comparing the reconstructed meshes directly against GT CAD models using three metrics: the 3D Volumetric Intersection over Union (IoU), which measures global spatial overlap; the Chamfer Distance (CD), which quantifies the average proximity between predicted and true point clouds; and the Hausdorff Distance (HD), which captures the maximum geometric deviation between the reconstructed and GT surfaces. To ensure accurate evaluation, each reconstructed model is normalized and rotated to align its coordinate frame with the canonical SPE3R GT frame. Together, these metrics provide complementary measures of reconstruction quality and ensure that the generated models are geometrically reliable for downstream pose estimation tasks.

\subsection{Pose Estimation Metrics}
From the Satellite Pose Estimation Challenge  \cite{chen2019satellite}, we consider the orientation error $e_{R}$, which is computed using the quaternion representation to avoid singularities and ensure numerical stability. The error is defined as the angular distance between the estimated quaternion $q_{est}$ and the ground truth $q_{gt}$:
\begin{equation}
e_{R} =
2\cos^{-1}(|q_{gt} \cdot q_{est}|)
\end{equation}
We evaluate the pointing accuracy of the translation vector, known as the Line-of-Sight (LOS) error. This metric decouples the directional accuracy from the absolute range estimation. The pointing error $e_{p}$ is calculated as the angular separation between the estimated and ground truth translation vectors:
\begin{equation}
e_{p} = \cos^{-1} \left( \frac{\mathbf{t}_{est} \cdot \mathbf{t}_{gt}}{\|\mathbf{t}_{est}\| \|\mathbf{t}_{gt}\|} \right)
\end{equation}
The inclusion of $e_{p}$ is related to the inherent limitations of monocular vision systems, i.e. the infeasibility to retrieve the absolute scale of a reconstructed object from a single monocular image without prior knowledge of the target's physical dimensions.
Absolute distance can generally only be recovered through triangulation, which requires multiple views and a known change in the chaser's position and velocity, or by assuming a fixed model scale during the 2D-3D matching process.  This limitation is also central to angles-only navigation for rendezvous and proximity operations, where range information is not instantaneously available and must instead become observable through the relative motion of the chaser and target \cite{aon}. Consequently, the translation error is often dominated by range inaccuracies, while $e_{p}$ provides a more direct measure of the geometric alignment performance.

To evaluate the geometric consistency of the estimated pose, we compute the Mean Reprojection Error (MRE) using the set of 2D-3D correspondences that survive the PnP RANSAC filtering. The MRE is defined as the average Euclidean distance, in pixels, between the observed 2D keypoints and the 3D points projected back onto the image plane using the estimated rotation $\mathbf{R}$, translation $\mathbf{t}$, and camera intrinsics $\mathbf{K}$. This metric quantifies the alignment accuracy of the PnP solver and serves as a measure of how well the reconstructed 3D model represents the visual features observed in the 2D image.

\subsection{Frame ambiguity} 

One of the challenging aspect of evaluating pose for reconstructed spacecraft models is that the object-fixed reference frame is not uniquely defined by the geometry alone. In monocular spacecraft shape-and-pose estimation, the estimated pose especially orientation is meaningful only relative to the body frame assigned to the reconstructed shape; however, when that frame is generated from structural rules such as aligning axes with the principal or largest dimensions of the object, the rule may constrain the axis ordering without uniquely determining the positive and negative directions of those axes. Dimension-based body-axis alignment cannot enforce consistent positive and negative axis directions across different spacecraft models, so a prediction may be penalized even when the estimated shape and pose are geometrically consistent with the image, but expressed in a different valid body-frame convention~\cite{bates2025removing}. This issue is further complicated by the fact that the reconstructed object is not a perfect replica of the GT model. Small geometric distortions, missing structures, smoothed components, or reconstruction artifacts can change the apparent principal dimensions of the shape, making frame extraction from the reconstruction less stable than frame extraction from the GT mesh. Thus, even if the same axis-assignment rule is applied to both models, the reconstructed frame may not align exactly with the GT frame before the sign ambiguity is considered. For a right-handed coordinate system, the valid sign-preserving frame conventions are the original frame and the three $180^\circ$ rotations about the body axes,
$(x,y,z)$, $(x,-y,-z)$, $(-x,y,-z)$, and $(-x,-y,z)$, corresponding to the permutation set used in the paper's ambiguity-aware rotation loss~\cite[Eq.~(8)-(9)]{bates2025removing}. This is related to, but distinct from, object symmetry: even when a spacecraft is not treated as physically symmetric, the constructed coordinate frame can still have an arbitrary sign convention. Consequently, a direct comparison between the estimated rotation and the GT rotation can report a large orientation error simply because the reconstructed body frame is flipped relative to the GT convention. To avoid treating this arbitrary convention mismatch as a physical pose error, we evaluate the rotation under each valid right-handed frame permutation and retain the convention that minimizes the pose error~\cite[Eq.~(9)]{bates2025removing}.

\subsection{FoundationPose}
We adopt NVIDIA’s FoundationPose \cite{foundationposewen2024} as a benchmark for our pipeline, since FoundationPose is a generalist 6D pose estimator and tracker trained on large-scale synthetic data from diverse 3D object repositories with domain randomization, demonstrating strong zero-shot generalization to unseen objects without fine-tuning. The method utilizes a transformer-based architecture to establish an iterative pose refinement process, where a neural network predicts the relative transformation required to align a 3D model with the observed image features. 

FoundationPose is a model-based 6D object pose estimation method that, for each pose, takes as input an RGB image, an object mask, a depth map, a normalized 3D mesh in OBJ format, and camera intrinsics. In our setup, all required inputs are provided by SP3ER except for the depth map. Since absolute depth information is not directly available in the normalized dataset, we generate it by rendering the GT 3D model from the known camera pose. Although this provides references with idealized depth information, it establishes a baseline to compare the underlying pose estimation logic. However, it must be noted that FoundationPose’s dependency on depth maps limits its feasibility for typical monocular space missions where active depth sensors are unavailable. 

\section{Experimental Design and Results} \label{sec:s4}
To quantify the performance and robustness of the DreamSat-Pose framework, we conduct a series of comparative simulations designed to isolate the impact of geometric fidelity on pose accuracy. Our experimental campaign is divided into two primary configurations. The first experiment is GT-based pose estimation. In this configuration, we provide both DreamSat-Pose and FoundationPose with GT meshes provided by the SPE3R dataset, to establish an approximate upper bound for pose estimation performance. This allows us to determine the maximum achievable accuracy of the matching and tracking algorithms when structural uncertainty is removed. The second experiment is reconstruction-based pose estimation. To evaluate the pipeline's utility for uncooperative targets, we perform pose estimation using the 3D meshes reconstructed directly from single-view SPE3R images. First, we consider only one reconstruction per test spacecraft. Then, due to reconstruction processing times, we select a subset of 100 images per spacecraft, split equally between black and Earth backgrounds (50 images each). These experiments serve as full-pipeline tests to determine whether a machine-learned geometric proxy can effectively substitute for a pre-existing CAD model in proximity operations. In DreamSat-Pose, the binary mask defines the object crop and restricts the extraction of image features to the spacecraft region. In FoundationPose, the same mask provides the object region required for pose initialization. For the reconstruction-based experiments, the masks are additionally used to remove the image background and prepare the isolated spacecraft image for single-view reconstruction. While the results presented are based on GT masks, preliminary experiments using the Segment Anything Model (SAM) \cite{kirillov2023segment} to automate foreground segmentation achieved comparable performance. This highlights the pipeline's potential for deployment in scenarios where manual mask annotations are unavailable.
Table \ref{tab:pose_results} summarizes the pose estimation performance for DreamSat-Pose and FoundationPose which are discussed in the following sections.  

\begin{table}[htbp]
\centering
\small
\setlength{\tabcolsep}{3pt}
\renewcommand{\arraystretch}{0.9}
\begin{tabular}{llrrrrrrrrr}
\toprule \midrule
\textbf{Model} & \textbf{Test} & \textbf{$N$} &
\multicolumn{4}{c}{\textbf{Orientation Error ($^\circ$)}} &
\multicolumn{4}{c}{\textbf{Pointing Error ($^\circ$)}} \\
\cmidrule(lr){4-7} \cmidrule(lr){8-11}
& & & Mean & Median & Min & Max & Mean & Median & Min & Max \\
\midrule
FoundationPose & GT Test & 7000 & 29.64 & 2.31 & 0.05 & 180.00 & 0.095 & 0.071 & 0.001 & 6.188 \\
FoundationPose & Recon Test & 7000 & 53.35 & 55.66 & 0.23 & 116.59 & 0.919 & 0.492 & 0.003 & 5.594 \\
FoundationPose & Recon 100per & 700 & 49.84 & 52.48 & 0.89 & 116.59 & 0.809 & 0.445 & 0.000 & 6.626 \\
DreamSat-Pose & Validation & 11395 & 29.32 & 12.77 & 0.27 & 118.08 & 0.092 & 0.050 & 0.000 & 3.547 \\
DreamSat-Pose & GT Test & 7000 & 48.90 & 47.65 & 0.57 & 110.88 & 0.133 & 0.094 & 0.001 & 2.637 \\
DreamSat-Pose & Recon Test & 7000 & 58.73 & 61.89 & 1.02 & 118.09 & 0.157 & 0.118 & 0.001 & 2.349 \\
DreamSat-Pose & Recon 100per & 700 & 57.84 & 59.30 & 2.91 & 114.93 & 0.179 & 0.123 & 0.004 & 2.139 \\
\midrule \bottomrule
\end{tabular}
\caption{Pose estimation performance for FoundationPose and DreamSat-Pose. Orientation and pointing errors are reported in degrees.}
\label{tab:pose_results}
\end{table}
\begin{figure}[H]
\centering
\begin{subfigure}[t]{0.32\textwidth}
    \centering
    \includegraphics[width=\linewidth]{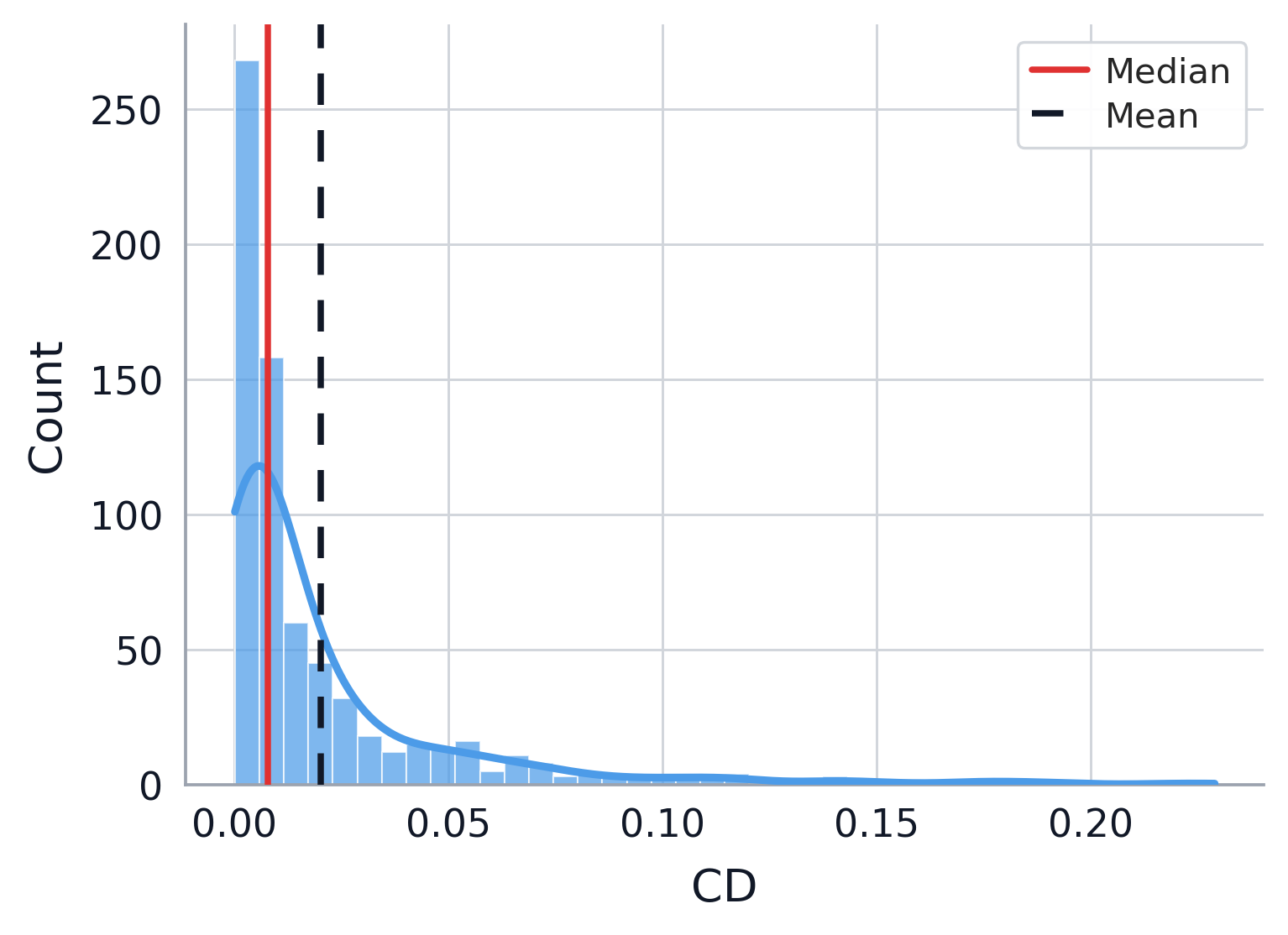}
    \caption{CD}
    \label{fig:reconstruction_cd}
\end{subfigure}
\hfill
\begin{subfigure}[t]{0.32\textwidth}
    \centering
    \includegraphics[width=\linewidth]{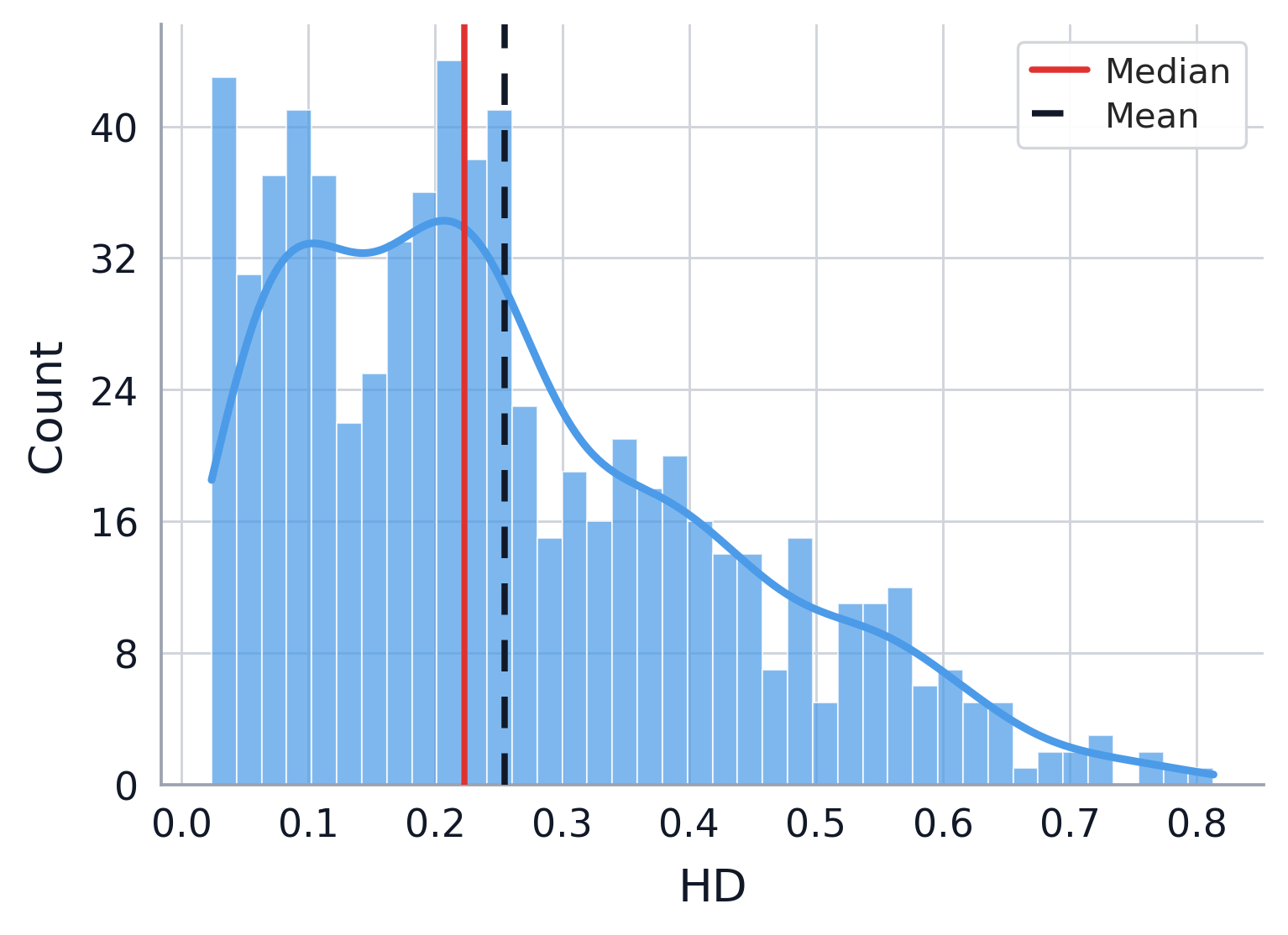}
    \caption{HD}
    \label{fig:reconstruction_hd}
\end{subfigure}
\hfill
\begin{subfigure}[t]{0.32\textwidth}
    \centering
    \includegraphics[width=\linewidth]{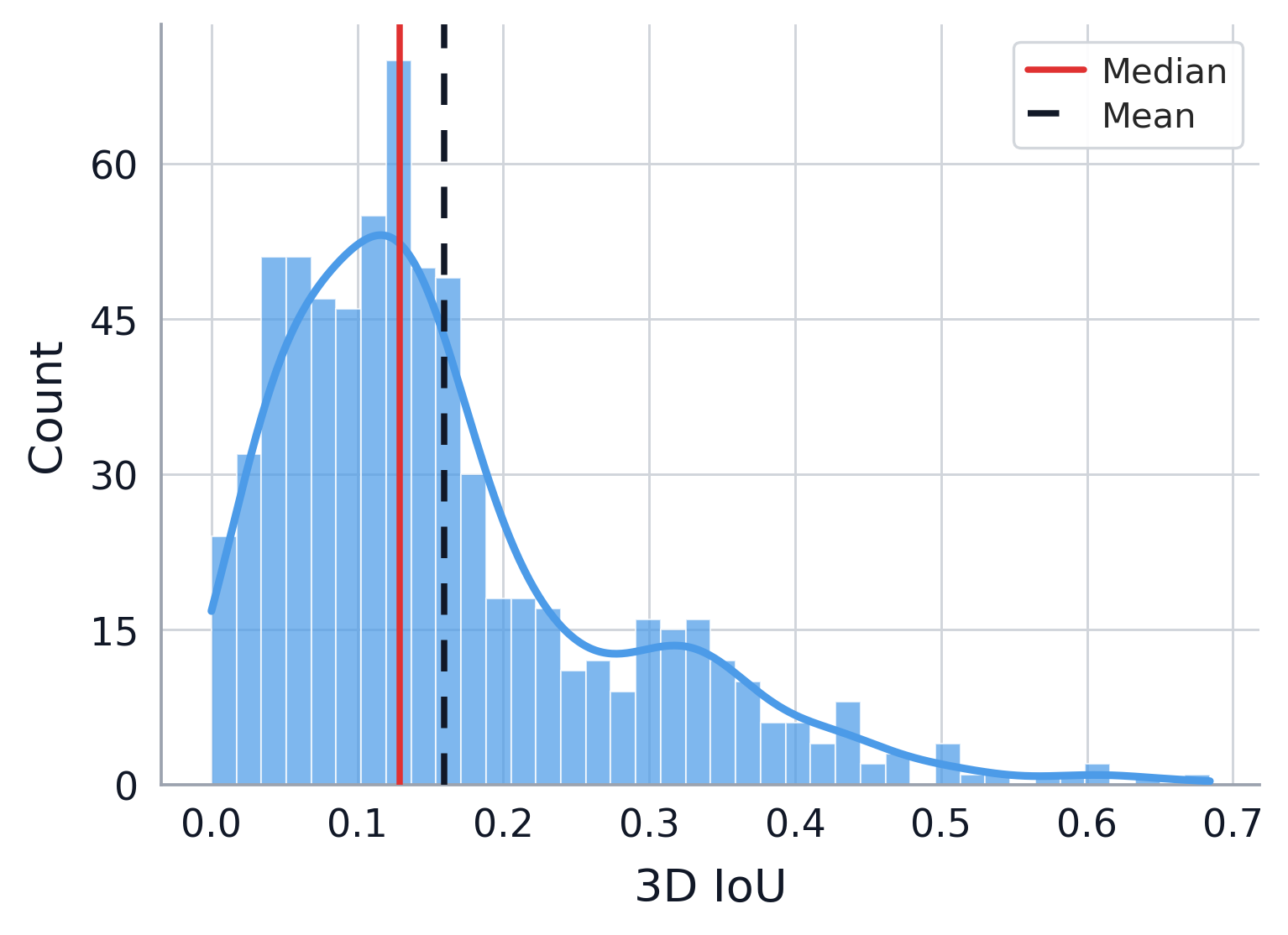}
    \caption{3D IoU}
    \label{fig:reconstruction_iou}
\end{subfigure}
\caption{Distribution of 3D reconstruction performance metrics across the evaluation dataset. Lower values indicate better performance for CD and HD, whereas values closer to one are better for IoU.}
\label{fig:reconstruction_metrics}
\end{figure}
Before evaluating pose estimation, the quality of the single-view reconstructions was assessed using the three-dimensional intersection over union (3D IoU), Hausdorff distance (HD), and Chamfer distance (CD), as shown in Fig.~\ref{fig:reconstruction_metrics}. The same reconstructed meshes were subsequently provided to both FoundationPose and DreamSat-Pose, ensuring that differences in pose-estimation performance reflect the robustness of each method to identical reconstruction errors.

\subsection{DreamSat-Pose}
\begin{figure}[H]
    \centering
    \includegraphics[width=0.70\textwidth]{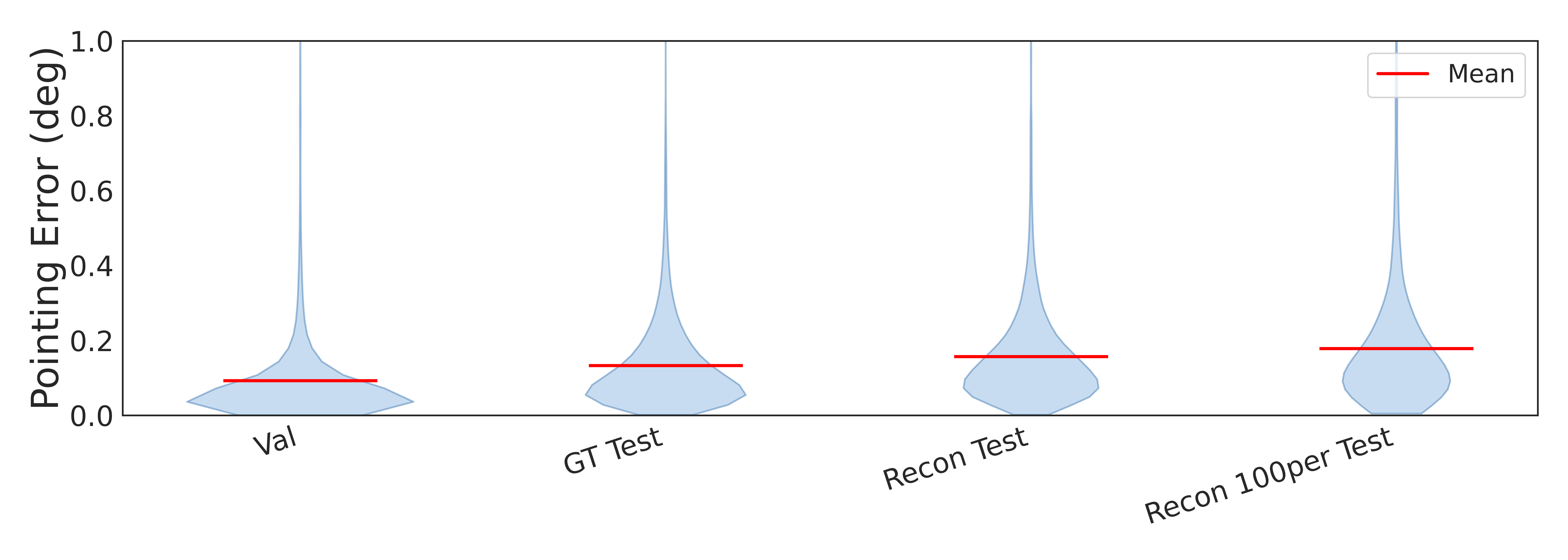}

    \caption{Pointing error distributions for DreamSat-Pose.}
    \label{fig:Dreamsat_dir_metric}
\end{figure}

\begin{figure}[H]
    \centering
    \includegraphics[width=0.70\textwidth]{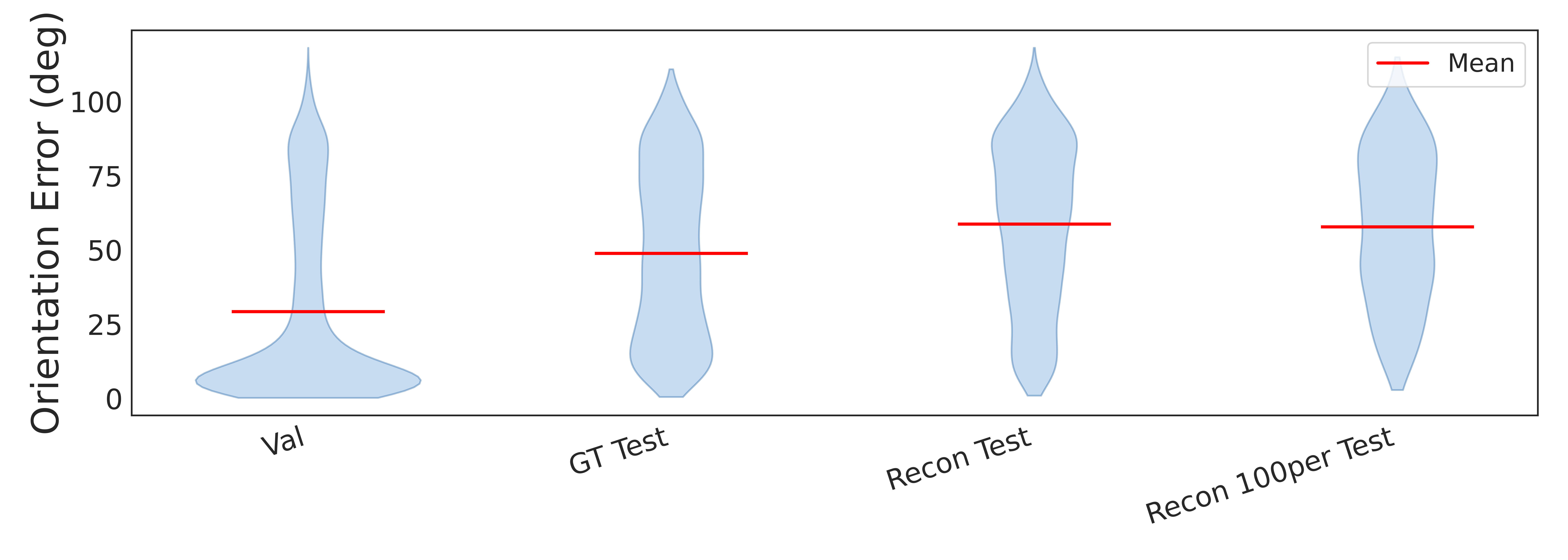}
    \caption{Orientation error distributions for DreamSat-Pose.}
    \label{fig:Dreamsat_ori_metric}
\end{figure}

In this section, the results of the experimental campaign for DreamSat-Pose are reported. Figures \ref{fig:Dreamsat_dir_metric} and \ref{fig:Dreamsat_ori_metric} illustrate the error distributions for both LOS direction and orientation for the validation set, the test set using GT CAD meshes (GT-Test), and the full pipeline using single-view reconstructed meshes (Recon-Test). The directional error demonstrates remarkable precision across all categories, with mean errors remaining well below $1^\circ$. This indicates that the pipeline is highly robust in identifying the target's relative bearing, regardless of whether a GT or a reconstructed mesh is utilized. The low variance in the GT-Test and Recon-Test configurations suggests that the 2D-3D correspondence matching successfully captures the geometric centroid of the spacecraft.
In contrast, the orientation error exhibits higher sensitivity to the fidelity of the 3D model. While the validation set achieves a mean orientation error of approximately $29.32^\circ$, the performance on the test set degrades when moving from GT meshes (mean $\approx 48.90^\circ$) to reconstructed meshes (mean $\approx 58.73^\circ$). 
The wider error distribution in the reconstruction-based configuration reflects structural uncertainties in the geometric proxies, particularly as no image enhancement is utilized in this study. Future integration of techniques such as super-resolution or denoising could improve the fidelity of the single-view 3D reconstructions and bridge the performance gap with GT CAD models. Nevertheless, these results confirm that machine-learned reconstructions serve as a functional proxy for pose estimation in uncooperative scenarios where no prior model is available. 

 To assess the fidelity of the single-view 3D reconstructions in the Recon-Test results, we evaluate the reconstruction metrics. As shown in Fig. \ref{fig:reconstruction_metrics}, Hunyuan-3D-2.0 achieves performance comparable to that reported in DreamSat \cite{dreamsat2}, despite a significant increase in data complexity. While the original study uses screenshots of isolated models, i.e. lacking complex illumination or background contexts, the current work employs realistic synthetic imagery from the SPE3R dataset. Specifically, across 700 reconstructions, the distributions of CD, HD, and 3D IoU indicate that the generated point clouds preserve overall spacecraft geometry with moderate structural fidelity despite the challenges of single-view reconstruction. However, not all outputs conform to a satisfactory reconstruction, suggesting the need to investigate the use of image processing techniques to enhance the visual quality of the inputs provided to the reconstruction model. Notably, no clear correlation is observed between reconstruction quality metrics and downstream pose estimation performance, as samples with lower reconstruction error do not consistently produce lower pointing or orientation errors. The MRE of the RANSAC inlier points depends on the chosen threshold, averaging 0.9, 2.4, and 3.7 pixels for the 2, 5, and 8 pixel Ransac thresholds, respectively.

\begin{figure}[H]
    \centering

    \begin{subfigure}[b]{0.70\linewidth}
        \centering
        \includegraphics[width=\linewidth]{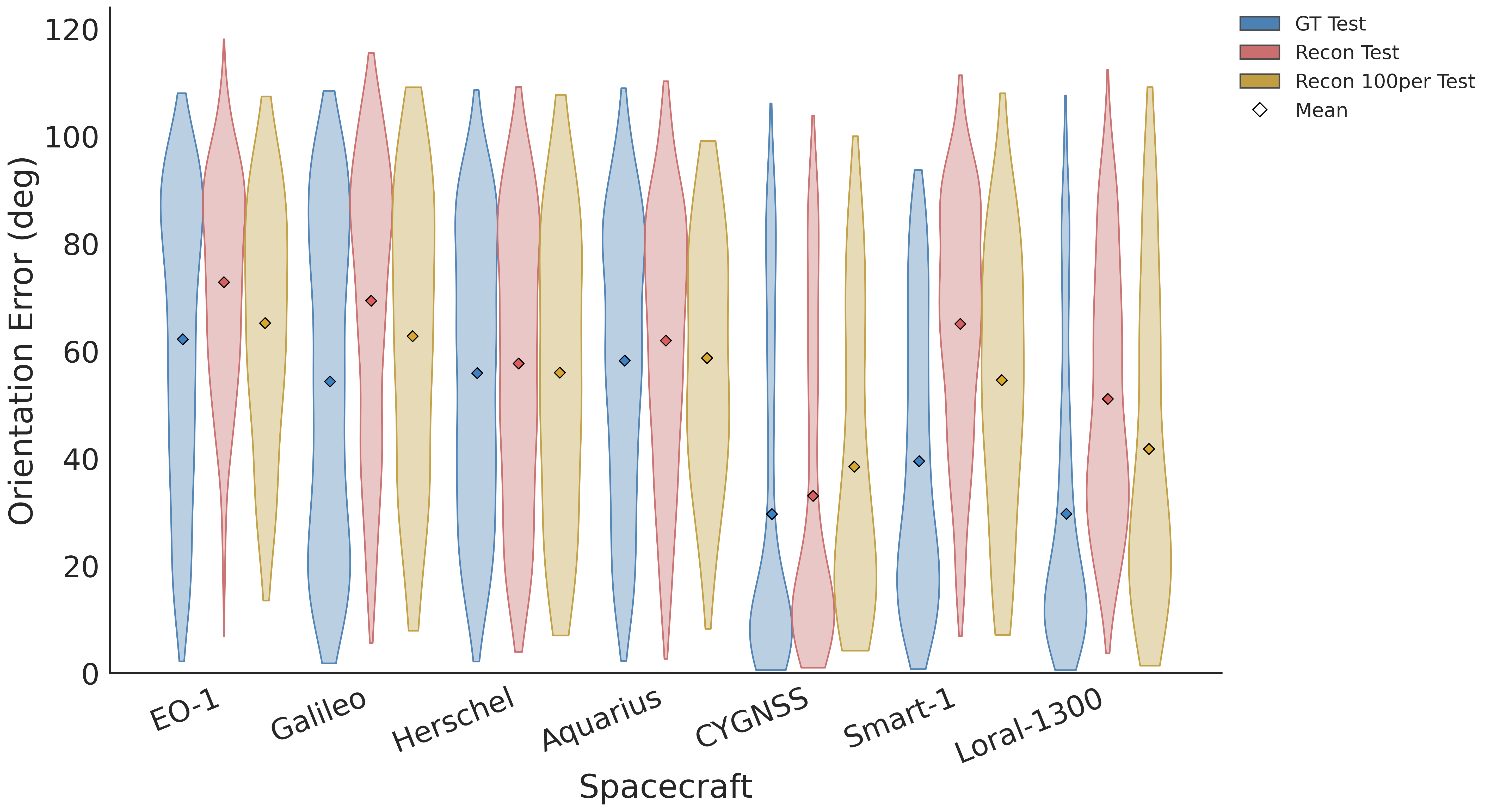}
        \caption{Per-spacecraft orientation error distributions.}
        \label{fig:dp_ori_perobj}
    \end{subfigure}

    \vspace{0.5cm}

    \begin{subfigure}[b]{0.70\linewidth}
        \centering
        \includegraphics[width=\linewidth]{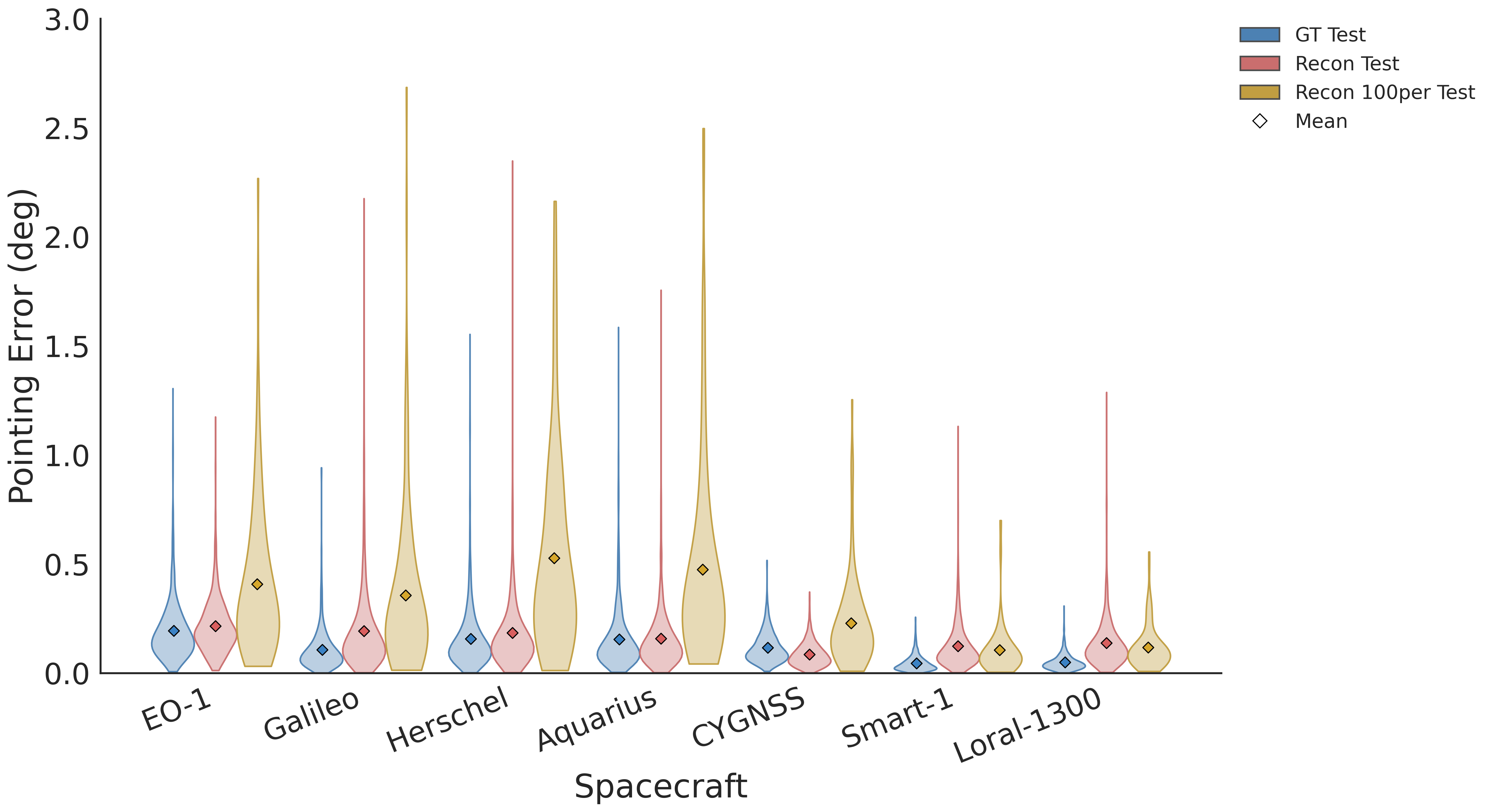}
        \caption{Per-spacecraft pointing-direction error distributions.}
        \label{fig:dp_dir_perobj}
    \end{subfigure}

    \caption{Per-spacecraft pose-estimation performance of the DreamSat-Pose pipeline under the GT Test, Recon Test, and Recon 100per Test conditions.}
    \label{fig:dp_perobj}
\end{figure}
Evaluation of the DreamSat-Pose pipeline for each spacecraft, as shown in Fig.~\ref{fig:dp_perobj}, demonstrates that the GT Test condition consistently outperforms the reconstruction-based conditions in orientation estimation. However, the magnitude of this performance gap varies considerably between spacecraft. Furthermore, the consistently low pointing errors across all conditions indicate that the pipeline can accurately estimate the direction towards the spacecraft, while the remaining error is primarily associated with rotation about this direction. These results suggest that reconstruction geometric inaccuracies contribute to pose degradation, but do not fully explain the large orientation errors, which are also likely influenced by ambiguous correspondences, repeated spacecraft structures, and inconsistencies between reconstructed and reference coordinate frames.

\subsection{Alternative Approach: FoundationPose}

In this section, the results of the experimental campaign for FoundationPose are reported. Similarly, FoundationPose is evaluated under two settings: a GT test setting, in which the method is supplied with the original SPE3R CAD meshes, and a reconstruction-based setting, in which the method uses meshes reconstructed from single-view images.

\begin{figure}[H]
    \centering
    \includegraphics[width=0.7\linewidth]{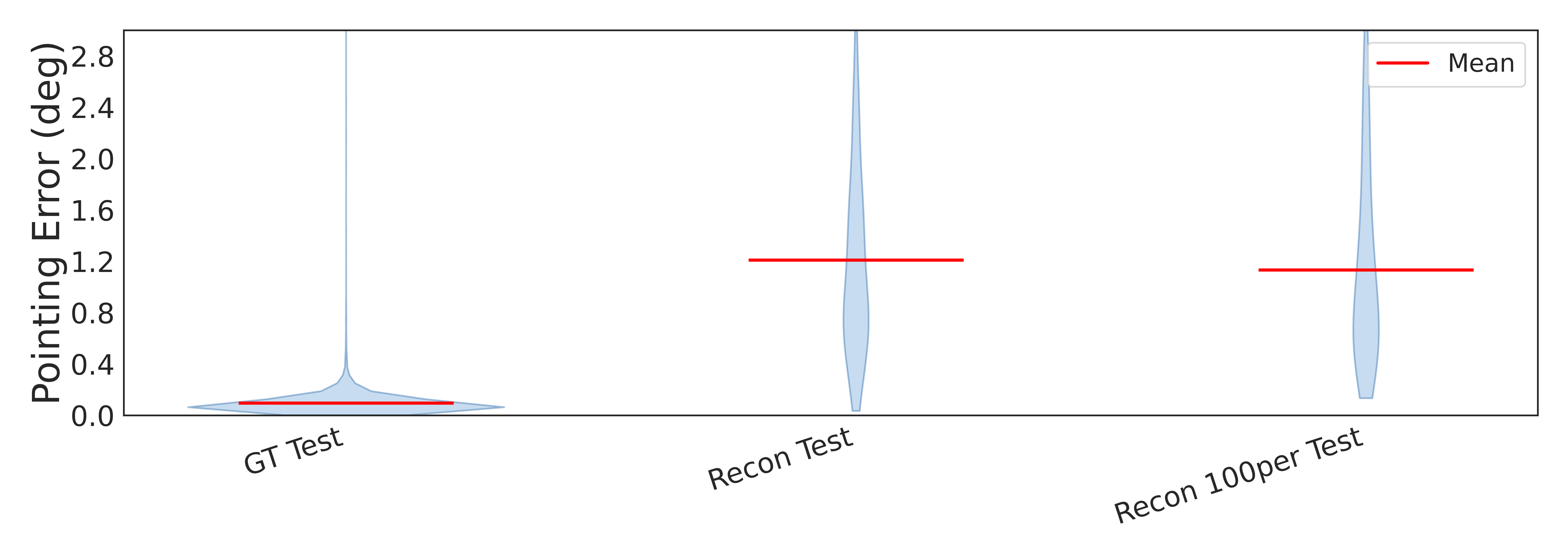}
    \caption{Pointing error distributions for FoundationPose across GT test and reconstructed test settings.}
    \label{fig:fp_dir_metric}
\end{figure}
\begin{figure}[H]
    \centering
    \includegraphics[width=0.7\linewidth]{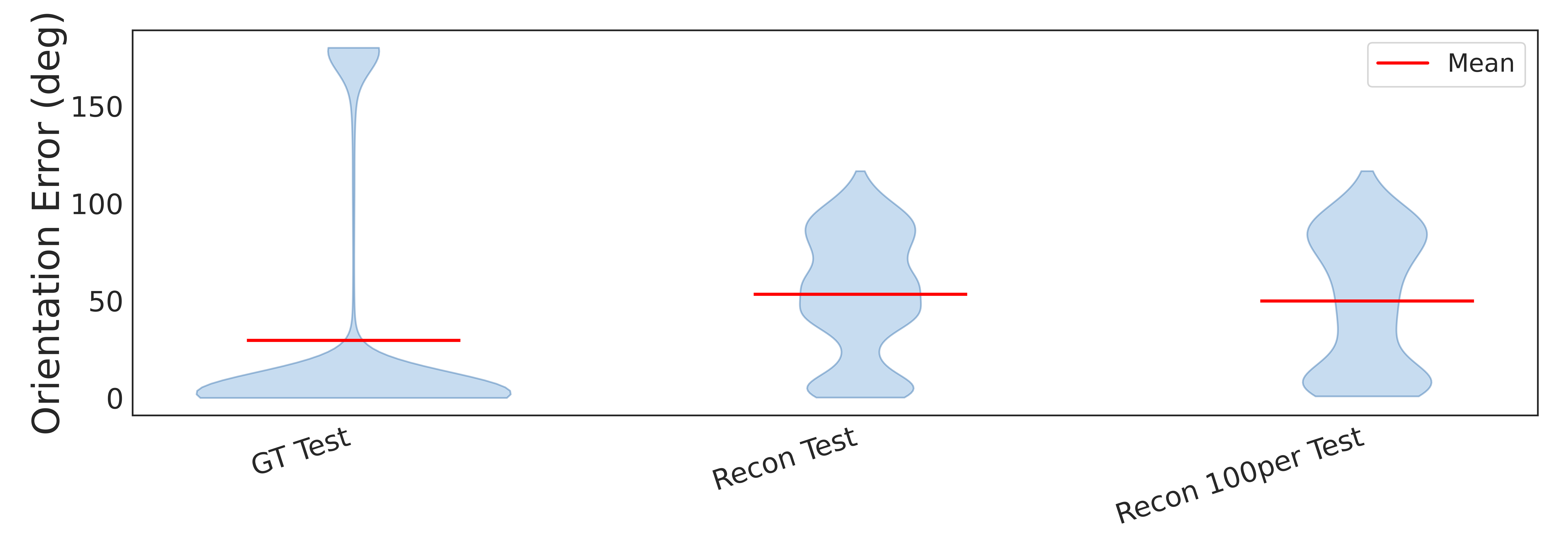}
    \caption{Orientation error distributions for FoundationPose after accounting for valid body-frame flips.}
    \label{fig:FP_perobj_amb}
\end{figure}
The pointing error distributions are shown in Fig.~\ref{fig:fp_dir_metric}. When supplied with the GT CAD models, FoundationPose achieves a mean pointing error of $0.095^\circ$ and a median error of $0.071^\circ$. When reconstructed meshes are used, the mean pointing error increases to $0.919^\circ$, while remaining below $1^\circ$. This degradation is expected because reconstructed meshes contain geometric imperfections, surface-smoothing artifacts, and structural inconsistencies that are absent from the GT models. Nevertheless, these results indicate that reconstructed geometry can still provide useful structural information for estimating the spacecraft pointing direction. It is important to note that FoundationPose was supplied with a depth map during inference. In this experimental setup, the depth map was generated from the known spacecraft geometry and camera pose, providing idealized geometric information that would generally not be available in practical monocular on-orbit scenarios. In real missions, depth sensing may be unavailable or degraded by sensor noise, sparse measurements, illumination constraints, or the absence of active depth sensors. Consequently, the FoundationPose results should be interpreted as an approximate upper bound under favorable sensing assumptions.
For orientation evaluation, a frame-ambiguity correction process was applied to account for valid right-handed body-frame flips between the reconstructed and GT models. This correction is necessary because geometrically equivalent models may use different axis conventions, resulting in large apparent orientation errors even when the estimated pose is geometrically consistent. After accounting for this ambiguity, FoundationPose achieves a mean orientation error of $53.35^\circ$ when using one reconstruction per $1{,}000$ images and $49.84^\circ$ when using $100$ reconstructions per object, as shown in Fig.~\ref{fig:FP_perobj_amb}. Thus, FoundationPose achieves an orientation error of approximately $50^\circ$ when evaluated using reconstructed geometry and ambiguity-aware frame alignment.

\begin{figure}[H]
    \centering

    \begin{subfigure}[b]{0.70\linewidth}
        \centering
        \includegraphics[width=\linewidth]{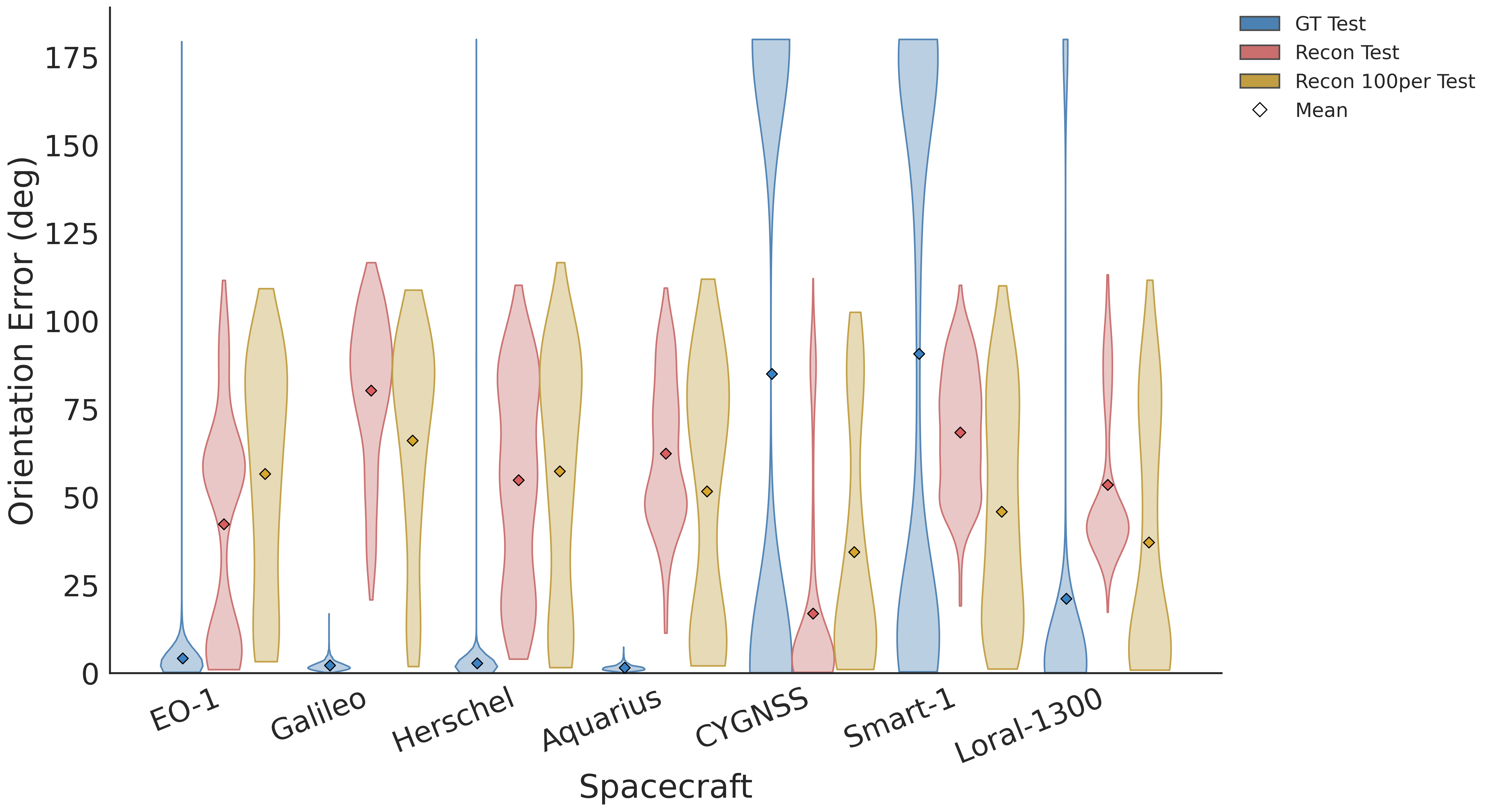}
        \caption{Per-spacecraft orientation error distributions.}
        \label{fig:fp_ori_perobj}
    \end{subfigure}

    \vspace{0.5cm}

    \begin{subfigure}[b]{0.70\linewidth}
        \centering
        \includegraphics[width=\linewidth]{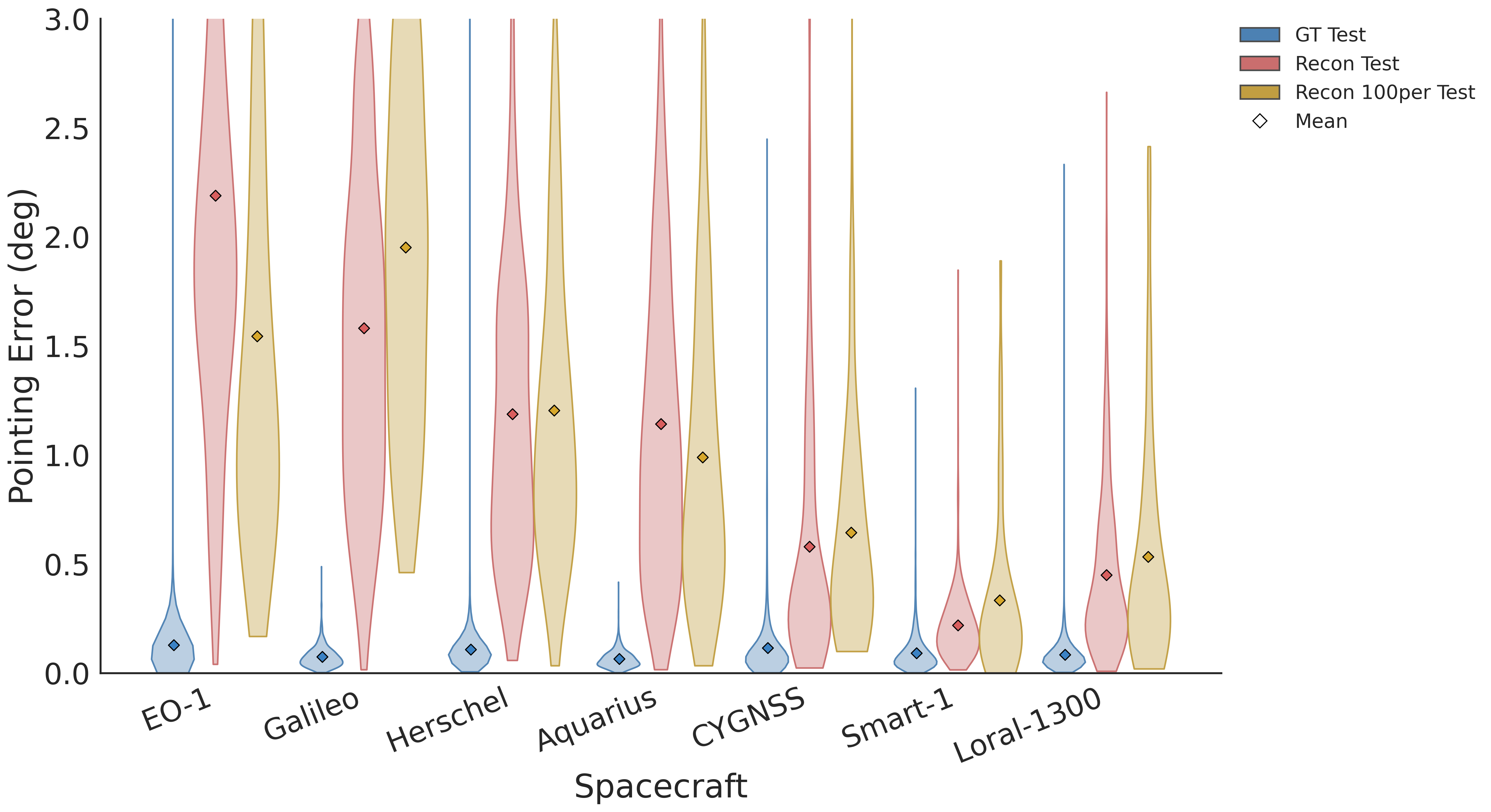}
        \caption{Per-spacecraft pointing-direction error distributions.}
        \label{fig:fp_dir_perobj}
    \end{subfigure}

    \caption{Per-spacecraft pose-estimation performance of FoundationPose under the GT Test, Recon Test, and Recon 100per Test conditions.}
    \label{fig:fp_pose_perobj}
\end{figure}

\subsection{FoundationPose: Effect of Spacecraft Symmetry} 
\begin{figure}[htbp]
    \centering
    \includegraphics[width=\textwidth]{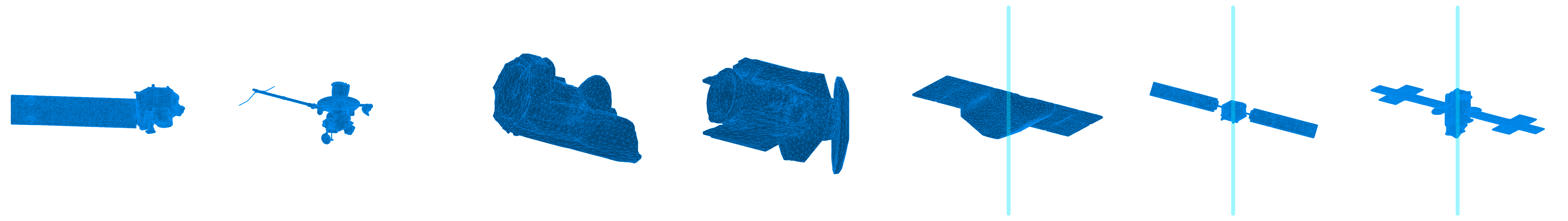}
    \caption{Representative spacecraft meshes projected in a common view: EO-1, Galileo, Herschel, Aquarius, CYGNSS, Smart-1, and Loral-1300. The vertical line indicates an approximate plane of symmetry where present.}
    \label{fig:symmetry_ordered_models}
\end{figure}

Although the overall pointing performance of FoundationPose remains strong, the orientation error distributions reveal a clear object-dependent behavior. In particular, Fig.~\ref{fig:FP_perobj_amb} shows that several samples exhibit large orientation errors despite accurate pointing estimates. This behavior is consistent with the inherent ambiguity of monocular pose estimation for geometrically symmetric spacecraft, where multiple rotations can produce nearly identical image projections. Figure~\ref{fig:fp_pose_perobj} shows that this ambiguity is highly object-dependent. FoundationPose achieves orientation errors below approximately $5^\circ$ for \textit{aquarius\_ud}, \textit{galileo\_actual\_05}, and \textit{herschel}, whose distinctive geometric structures uniquely constrain their orientation. In contrast, \textit{cygnss\_solo\_39}, \textit{eo-1\_final}, \textit{loral-1300com-main}, and \textit{smart\_1} exhibit substantially larger orientation errors despite maintaining accurate pointing estimates. This trend is further illustrated in Fig.~\ref{fig:symmetry_ordered_models}, which compares representative spacecraft meshes in a common view. Spacecraft with distinctive, asymmetric structures provide stronger geometric constraints for rotation recovery. Conversely, spacecraft with repeated components or approximate planes of symmetry can admit multiple visually plausible orientations. In these cases, FoundationPose may predict an orientation that is visually similar to the ground truth but differs by a rotation about a symmetry axis. Therefore, the reported orientation error may overestimate the practical localization error for highly symmetric spacecraft. The consistently low pointing errors nevertheless indicate that FoundationPose recovers the spacecraft line of sight accurately, even when orientation remains ambiguous.

\subsection{Runtime and Memory Performance}
Once initialized, the DreamSat-Pose runtime is about 0.034 seconds per image and includes 2D feature extraction using DINOv3, 3D feature computation using the DGCNN representation, transformer-based correspondence matching, and final pose estimation using EPnP. The maximum system memory consumption during end-to-end inference is approximately 1.62 GB of RAM. The 3D reconstruction takes about 35-40 seconds when using the Hunyuan-3D-2.0 model. However, a faster option is available, Hunyuan-3D-2.0-mini-turbo-flash, which takes about 4 seconds per image; its usage and a trade off between reconstruction quality and speed will be explored in future work. On the other hand, FoundationPose runtime is approximately 1 second per image.

\section{Conclusion}

This paper proposes DreamSat-Pose, a framework for 6-DoF spacecraft pose estimation that operates under the minimal assumptions of a single RGB image without requiring depth sensing or a-priori CAD models. By reformulating pose estimation as a correspondence problem between 2D image features and generative 3D reconstructions, the pipeline provides a scalable solution for navigating around uncooperative or unknown targets. The results from our experiments demonstrate that high-quality 3D reconstructions can effectively serve as proxies for GT geometry. Moreover, the system demonstrates strong generalization to unseen spacecraft, suggesting that the learned representations capture fundamental geometric structures rather than object-specific appearance. At the same time, the results show that orientation remains the most challenging component of the pose estimation task. While the pipeline achieves consistently accurate pointing estimates, the orientation error remains more sensitive to reconstruction artifacts, object symmetries, repeated structural patterns, and body-frame ambiguities. The strong performance observed when using GT meshes indicates that the learned 2D-3D correspondence and PnP-based pose estimation pipeline can recover accurate pose information when reliable geometry is available. Therefore, as single-view reconstruction outputs continue to improve, the reconstruction-based results suggest a clear possibility for using generated 3D models as practical geometric proxies in monocular spacecraft pose estimation. The significance of this work to the future of autonomous space operations is twofold. First, it enables autonomous missions to unprepared and unknown targets by bypassing the current bottleneck where navigation systems rely on GT CAD models or cooperative markers; data that is often lost or outdated for defunct objects and debris. Second, it offers a path toward a significant reduction in sensor size, weight, and power (SWaP) by demonstrating that accurate 6-DoF estimation can be achieved through a monocular pipeline rather than relying on heavy, power-hungry LiDAR or stereo-camera rigs. Despite these promising outcomes, several limitations remain for operational deployment. The current end-to-end inference latency can be suitable for situational awareness but might exceed the requirements for real-time closed-loop control and autonomous docking. Furthermore, orientation estimation remains a challenge due to the inherent geometric symmetries and repeated structural patterns of many spacecraft. Another limitation is the resolution and framing of the input imagery. The SPE3R images are $256 \times 256$ pixels, and the spacecraft does not always occupy the full image. Instead, the object of interest is often embedded within the frame, and for distant targets, the visible satellite can occupy only a small number of pixels. This limits the amount of visual detail available to the feature extractor and makes it harder to obtain meaningful image embeddings and reliable 2D-3D correspondences. Also, the reliance on photorealistic renderings leaves a realism gap, as the model has yet to be challenged by the sensor noise, compression artifacts, and extreme specular reflections characteristic of actual flight imagery. Future work will focus on investigating model distillation and weight quantization to reduce the computational footprint and memory usage, multi-hypothesis pose estimation and uncertainty modeling to better capture symmetric configurations, and validating the pipeline against realistic sensor noise and motion blur. Future work should also explore higher-resolution inputs, improved cropping and upscaling strategies, and image enhancement techniques such as super-resolution and denoising to improve feature extraction when the spacecraft occupies only a small region of the image. Ultimately, DreamSat-Pose aims to provide a robust framework for autonomous interaction with the complex and unpredictable modern orbital environment.

\section*{Code Availability} The code for this work can be accessed on GitHub (\url{https://dreamsatpose.github.io/}).

\section*{Acknowledgements}
From the MIT side, the research was sponsored by the Department of the Air Force Artificial Intelligence Accelerator and was accomplished under Cooperative Agreement Number FA8750-19-2-1000. The views and conclusions contained in this document are those of the authors and should not be interpreted as representing the official policies, either expressed or implied, of the Department of the Air Force or the U.S. Government. The U.S. Government is authorized to reproduce and distribute reprints for Government purposes notwithstanding any copyright notation herein. Paolo Panicucci is supported by the Progetto Rocca Fellowship.
%



\bibliographystyle{IEEEtran}
\bibliography{main}

\end{document}